\documentclass[acmsmall,screen]{acmart}

\usepackage{latexsym}
\usepackage{graphicx}
\usepackage{subfigure}
\usepackage{microtype}
\usepackage{booktabs}
\usepackage{multirow}
\usepackage{multicol}
\usepackage{makecell}
\usepackage{enumitem}

\AtBeginDocument{%
  \providecommand\BibTeX{{%
    \normalfont B\kern-0.5em{\scshape i\kern-0.25em b}\kern-0.8em\TeX}}}

\begin{document}

\title{Target-constrained Bidirectional Planning for Generation of Target-oriented Proactive Dialogue}

\author{Jian Wang}
\affiliation{%
  \institution{The Hong Kong Polytechnic University}
  \city{Hong Kong}
  \country{China}
}
\email{jian-dylan.wang@connect.polyu.hk}

\author{Dongding Lin}
\affiliation{%
  \institution{The Hong Kong Polytechnic University}
  \city{Hong Kong}
  \country{China}
}
\email{dongding88.lin@connect.polyu.hk}

\author{Wenjie Li}
\affiliation{%
  \institution{The Hong Kong Polytechnic University}
  \city{Hong Kong}
  \country{China}
}
\email{cswjli@comp.polyu.edu.hk}

\renewcommand{\shortauthors}{Wang et al.}

\begin{abstract}
Target-oriented proactive dialogue systems aim to lead conversations from a dialogue context toward a pre-determined target, such as making recommendations on designated items or introducing new specific topics. To this end, it is critical for such dialogue systems to plan reasonable actions to drive the conversation proactively, and meanwhile, to plan appropriate topics to move the conversation forward to the target topic smoothly. In this work, we mainly focus on effective dialogue planning for target-oriented dialogue generation. Inspired by decision-making theories in cognitive science, we propose a novel target-constrained bidirectional planning (TRIP) approach, which plans an appropriate dialogue path by looking ahead and looking back. By formulating the planning as a generation task, our TRIP bidirectionally generates a dialogue path consisting of a sequence of <action, topic> pairs using two Transformer decoders. They are expected to supervise each other and converge on consistent actions and topics by minimizing the decision gap and contrastive generation of targets. Moreover, we propose a target-constrained decoding algorithm with a bidirectional agreement to better control the planning process. Subsequently, we adopt the planned dialogue paths to guide dialogue generation in a pipeline manner, where we explore two variants: prompt-based generation and plan-controlled generation. Extensive experiments are conducted on two challenging dialogue datasets, which are re-purposed for exploring target-oriented dialogue. Our automatic and human evaluations demonstrate that the proposed methods significantly outperform various baseline models.

\end{abstract}

\setcopyright{acmcopyright}
\acmJournal{TOIS}
\acmYear{2024} 
\acmVolume{1} 
\acmNumber{1} 
\acmArticle{1} 
\acmMonth{3} 
\acmPrice{15.00}
\acmDOI{XXXXXXX.XXXXXXX}

\ccsdesc[500]{Computing methodologies~Discourse, dialogue and pragmatics}
\ccsdesc[500]{Computing methodologies~Natural language generation}
\keywords{Target-oriented dialogue, Dialogue generation, Bidirectional planning}

\maketitle

\section{Introduction}
Human-machine dialogue systems have made significant progress in chatting with users for entertainment, e.g., open-domain dialogues \cite{zhang-etal-2020-dialogpt,wang2020improving,huang2020challenges}, and assisting users in accomplishing specific tasks, e.g., task-oriented dialogues \cite{madotto2018mem2seq,wu2019global,wang-etal-2020-dual}. Despite passively responding to users, dialogue systems can also take a more proactive role \cite{wu-etal-2019-proactive,deng2023survey} to introduce new interesting topics to users. Such a target-oriented proactive dialogue system looks more intelligent, sociable, and capable of directing the users towards topic areas that the system knows how to talk about \cite{gupta-etal-2022-target}. However, previous studies \cite{tang-etal-2019-target,wu-etal-2019-proactive,sevegnani-etal-2021-otters,gupta-etal-2022-target} mainly focus on the scenario of open-domain dialogues. They define the target as a commonsense topic and explore bridging an initial dialogue context and the given topic. Such a scenario is difficult to be generalized to real-world applications.

\begin{figure*}[t!]
\centering
\includegraphics[width=0.98\textwidth]{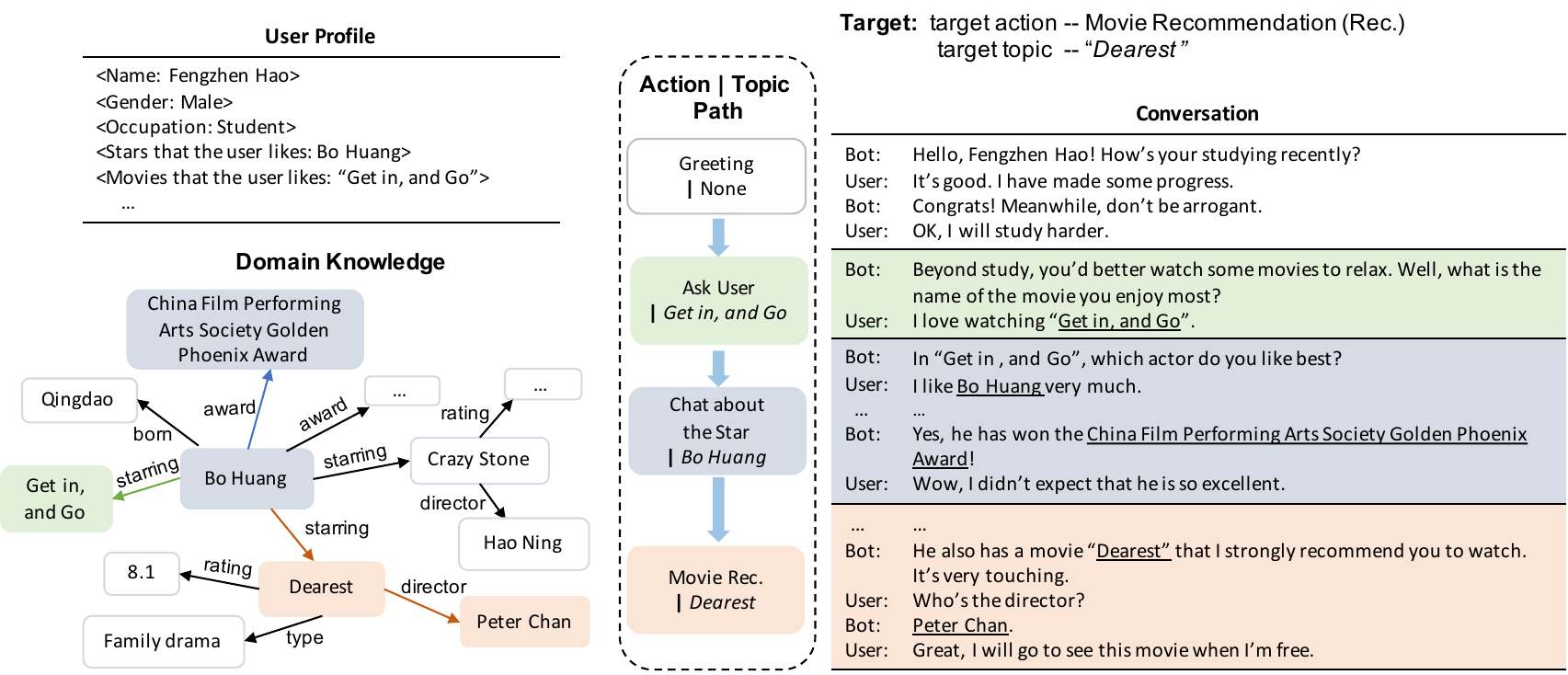}
\caption{An illustrative example from the re-purposed DuRecDial \cite{liu-etal-2020-towards-conversational} dataset. Given a pre-determined target and a dialogue context, our objective is to generate utterances that proactively and smoothly lead the conversation to achieve the target.}
\label{fig:example}
\end{figure*}

In this work, we take a further step towards a more challenging target-oriented dialogue scenario, where the target is defined as an <action, topic> pair, such as providing recommendations for a specific topic that possibly attracts users. It requires the system to take more engaging actions to achieve the target, such as social chitchat, user exploration, topic elicitation, recommendation, etc. As an example shown in Figure \ref{fig:example}, suppose there is an explicit target, i.e., to recommend a specific movie named ``\textit{Dearest}'', the system (i.e., Bot) is required to lead the conversation (e.g., ``\textit{greeting}'' $\rightarrow$ ``\textit{ask user}'' $\rightarrow$ ``\textit{chat about the star}'' $\rightarrow$ ``\textit{movie recommendation}'') so as to recommend the target movie when appropriate. It needs to consider the pre-determined target, dialogue history, and grounding domain knowledge (and user profile, if any). Particularly, the grounding domain knowledge associated with domain-specific topics and relevant attributes, is crucial to enable multiple topic transitions (e.g., warm-up chitchat $\rightarrow$ ``\textit{Get in, and Go}'' $\rightarrow$ ``\textit{Bo Huang}'' $\rightarrow$ ``\textit{Dearest}''). It is non-trivial to solve target-oriented dialogue generation for two reasons: (1) The system needs to keep the conversation engaging and proactively drive the conversation; (2) The system is desired to move the conversation forward to the target topic coherently and arouse the user's interest in the target topic to be recommended.

To address the above challenges, we observe that effective dialogue planning \cite{wang2022follow,wang2023target} is essential for target-oriented dialogue generation. In order to achieve its target, the system needs to plan reasonable actions and appropriate topics to smoothly move the conversation forward to the target topic before generating each system utterance\footnote{We use the term ``system utterance'' in this paper rather than ``response'' used in a lot of related work since the system needs to proactively lead the conversation in most cases.}. 
According to decision-making theories \cite{triantaphyllou2000multi,hunt2021formalizing} in cognitive science, humans tend to look ahead (forward) and look back (backward) when making decisions to achieve a long-term goal. Such bidirectional thinking alleviates short-sighted cognition and drives people to think about the complete decision path more. 
Similarly, in target-oriented proactive dialogue, the target has been designated in advance and should be bounded at the end of the dialogue path to be planned, backward path planning is effective in leveraging target-side information but insensitive to the coherence of the dialogue context. In contrast, forward path planning is more effective in generating a starting path point that is coherent with the dialogue context, while lacking the target-driven ability to enable the target to be bounded at the end of the dialogue path to be planned. 
With this in mind, we propose a \textbf{T}a\textbf{R}get-constrained b\textbf{I}directional \textbf{P}lanning (\textbf{TRIP}) method. The key point is to plan dialogue paths from both look-ahead (i.e., present-to-target) and look-back (i.e., target-to-present) directions. Generally, it is more appropriate when the look-ahead decision path is consistent with the look-back decision path. By formulating the planning as a generation task, our TRIP bidirectionally generates dialogue paths consisting of a sequence of <action, topic> pairs (see Figure \ref{fig:example}) based on an encoder-decoder architecture. Concretely, we first take widely-used pre-trained language models, e.g., BERT \cite{devlin-etal-2019-bert}, to encode complex input texts efficiently. Then, we employ two individual Transformer \cite{vaswani2017attention} decoders for dialogue path generation, with one to generate a dialogue path in the target-to-present direction and the other to generate one in the present-to-target direction. By minimizing the decision gap between the two directions, the two decoders are expected to provide supervision to each other and converge on a consistent dialogue path. In addition, we propose a contrastive generation mechanism (see Section \ref{para:contgen}) to enhance TRIP with the ability to better distinguish between the given target and non-targets. It enables TRIP to be more robust in generating the necessary target in the planned dialogue path accordingly. During inference, we propose a target-constrained decoding algorithm (see Section \ref{sec:decoding}) with a bidirectional agreement, which reduces the gap between inference and training and facilitates the model to generate an appropriate dialogue path as the ultimate output. 

Since each planned dialogue path outlines how to achieve the pre-determined target step by step, it is expected to help a dialogue system distill necessary knowledge and steer the system to generate more proper utterances with control. We adopt the planned dialogue path to guide dialogue generation in a pipeline manner, where we explore two variants: prompt-based generation (see Section \ref{sec:promptgen}) and plan-controlled generation (see Section \ref{sec:controlgen}). As part of this work, we re-purpose two existing recommendation-oriented dialogue datasets, namely DuRecDial \cite{liu-etal-2020-towards-conversational} and DuRecDial 2.0 \cite{liu2021durecdial}, for target-oriented dialogue generation through automatic target construction.
Extensive experiments are conducted to show the effect of planning and the performance of different dialogue generation methods. Both automatic and human evaluations demonstrate that our proposed methods significantly outperform various baseline models.

Overall, our contributions are summarized as follows:
\begin{itemize}
\item We introduce the target-oriented dialogue generation task and discuss its relation and difference (Section \ref{sec:relwork} and \ref{sec:prel}) compared to existing works.
\item We propose a novel target-constrained bidirectional planning (TRIP) approach (Section \ref{sec:trip}) for target-oriented proactive dialogue systems. Our TRIP plans a dialogue path consisting of a sequence of <action, topic> pairs that outline how to achieve the designated target proactively and smoothly.
\item We investigate both the prompt-based and plan-controlled methods (Section \ref{sec:pedg}) to leverage planned dialogue paths to guide dialogue generation effectively.
\item Experimental results show that our method achieves state-of-the-art performance in both automatic and human evaluations. Our extensive analysis provides some new insights into how planning affects target-oriented dialogue generation. 
\end{itemize}

\section{Related Work}
\label{sec:relwork}
Our work is mainly related to dialogue systems and content planning. We briefly review related work and clarify key differences compared with our work as follows.

\subsection{Target-oriented Dialogue}
Target-oriented dialogue systems work on the task of generating responses guided by the given target. According to the variety of the target, previous works have mainly focused on using a keyword \cite{tang-etal-2019-target,qin2020dynamic,zhong2021keyword}, a topic \cite{wu-etal-2019-proactive,sevegnani-etal-2021-otters}, and a concept or a sentence \cite{gupta-etal-2022-target} as the guided target.
For example, \cite{tang-etal-2019-target} introduced some coarse-grained keywords to control the intended content of the responses in open-domain dialogues, while \cite{zhong2021keyword} leveraged external commonsense knowledge graphs for keyword transitions. As a follow-up study, steering a dialogue towards a given keyword, or dialogue strategy learning, has also been explored in past work, including graph grounded policies \cite{xu-etal-2020-conversational,xu2020knowledge} and conversational lines \cite{ghazarian-etal-2021-discol}. For topic-guided dialogues, \cite{wu-etal-2019-proactive} investigated using an entity over a factual knowledge graph as the target topic, which requires the system to achieve a smooth transition from an initial topic to the given target topic. A new dataset called OTTers \cite{sevegnani-etal-2021-otters} was collected to explore one-turn topic transitions for open-domain response generation. More recently, \cite{gupta-etal-2022-target} proposed to identify a bridging path of commonsense knowledge concepts between the dialogue context and the target sentence using data augmentation. Our work is more related to prior settings \cite{lin2021target} on target topics and target sentences. However, existing works mainly focus on the scenario of open-domain target-guided dialogue, where they mainly consider guiding chitchat conversations to the target with transitions on commonsense topics. In comparison, we work on a more challenging setup that aims to achieve the target action for a designated target topic. It requires the system to take more engaging dialogue actions, such as social chitchat, user exploration, topic elicitation, and recommendation, to attract users so as to complete the target. We also clarify that existing studies on goal-oriented dialogue \cite{gonzalez2019retrieval,snell2022context} focus on the user-side goal or task, while our work explores the system-side target (or a specific goal).

\subsection{Recommendation-oriented Dialogue}
As a special type of task-oriented dialogue system, a recommendation-oriented dialogue system is desired to make recommendations through natural conversations with users.
It was the emergence of various recommendation-oriented dialogue datasets that helps push forward the research in this area, such as \textsc{GoRecDial} \cite{kang-etal-2019-recommendation}, TG-ReDial \cite{zhou-etal-2020-towards}, \textsc{INSPIRED} \cite{hayati-etal-2020-inspired}, and DuRecDial \cite{liu-etal-2020-towards-conversational}. As follow-up studies, CR-Walker \cite{ma-etal-2021-cr} was proposed to perform tree-structured reasoning over knowledge graphs, which can then be mapped into hierarchical dialogue acts to guide both item and response generations. MGCG \cite{liu-etal-2020-towards-conversational} and KERS \cite{zhang-etal-2021-kers} explored the transition policy from a non-recommendation dialogue to a recommendation-oriented one. 
There is another similar research area called conversational recommender systems (CRS) \cite{li2018towards,sun2018conversational,lin2023cola}.  Compared with recommendation-oriented dialogue systems, the main task of CRS lies in discovering user preferences \cite{xu-etal-2020-user,xu2021adapting}, asking clarifying questions about item attributes \cite{lei2020estimation,li2021deux}, and searching for optimal candidate items \cite{zhou2020improving,zhang2021kecrs,liang-etal-2021-learning}. In addition, \cite{deng2023unified} unified item recommendation and response generation into the same sequence-to-sequence (Seq2Seq) paradigm using prompt-based learning.
Nonetheless, most existing systems passively respond to a user, where they provide recommendations according to the user's expressed interests or requirements. Our work aims to endow a dialogue system with a more proactive role that can attract the user's interests and naturally lead user-engaged dialogues to achieve a pre-determined target.

\subsection{Content Planning for Language Generation}
There is a line of work \cite{puduppully2019data,hua2019sentence,moryossef2019step,su2021plan} that separates natural language generation into \textit{content planning} and \textit{surface realization}. Content planning mainly focuses on selecting the key contents (e.g., key phrases and entities) and arranging their orderings \cite{puduppully2019data,moryossef2019step}, followed by a neural generation stage that focuses only on realization. Different strategies have been explored for content planning. For example, \cite{shao2019long} proposed a hierarchical variational model for planning-based data-to-text generation, where a global latent variable models the diversity of planning and a sequence of local latent variables controls sentence realization. \cite{hua2020pair} presented a planning framework with iterative refinement to leverage large pre-trained language models for argument generation and article writing. For long-form text generation tasks, several studies \cite{hua2021dyploc,hu2022planet} conducted dynamic content planning while generating the output based on mixed language models to bridge the gap between content planning and sentence realization. Compared to these prior studies, our work is more related to planning for dialogue generation \cite{yarats2018hierarchical,wang2022follow}. We aim to address a more challenging dialogue generation task, where we propose a novel target-constrained bidirectional planning method to guide pre-trained language models to generate dialogue utterances more effectively.

\section{Preliminaries}
\label{sec:prel}
In this section, we aim to provide preliminaries about the problem formulation and introduce essential sub-tasks accordingly. Then, we briefly introduce our proposed method with respect to addressing the problem effectively.

Suppose we have a target-oriented dialogue corpus $\mathcal{D}=\{(\mathcal{K}_{i},\mathcal{P}_{i},\mathcal{H}_{i})\}_{i=1}^{N}$, where $N$ denotes the number of dialogue samples. $\mathcal{K}_{i}=\{k_{i,j}\}_{j=1}^{N_K}$ denotes a set of domain knowledge facts relevant to $i$-th dialogue with each element $k_{i,j}$ in form of a $\langle\textit{subject, relation, object}\rangle$ triple. $\mathcal{H}_{i}=\{(X_{i,t},Y_{i,t})\}_{t=1}^{T}$ denotes dialogue content with a total number of $T$ turns. $\mathcal{P}_{i}=\{(a_{i,l}, z_{i,l})\}_{l=1}^{L}$ denotes an annotated dialogue path for $i$-th dialogue, each path span specifies an action-topic pair (a dialogue action $a_{i,l}$ and a dialogue topic $z_{i,l}$). $ L$ is the number of unique action-topic pairs. Here, the dialogue topics are mainly constructed upon the domain knowledge $\mathcal{K}_{i}$, and each action/topic may affect multiple turns of dialogue. In some scenarios, there also exists a user profile $\mathcal{U}_{i}$ grounded on the $i$-th dialogue, which can be personal attributes or certain preferences.

Given a target $\mathcal{G}^{'}=(a_{T^{'}},z_{T^{'}})$ consisting of a target action $a_{T^{'}}$ and a target topic $z_{T^{'}}$, a dialogue history $\mathcal{H}^{'}$, and a set of relevant domain knowledge $\mathcal{K}^{'}$ (and a user profile $\mathcal{U}^{'}$, if any), our objective is to generate coherent utterances to engage the user in the conversation so as to achieve the target $\mathcal{G}^{'}$ when appropriate. Due to the complexity of the problem, it can be decomposed into three sub-tasks: (1) \textbf{action planning}, i.e., plan actions to determine where the conversation should go to lead the conversation proactively; (2) \textbf{topic planning}, i.e., plan appropriate topics to move forward to the target topic smoothly; (3) \textbf{dialogue generation}, i.e., generate an appropriate utterance to achieve the planned action and topic at each turn.

To address the above tasks, we propose a target-constrained bidirectional planning method to guide dialogue generation in a pipeline manner. The target-constrained bidirectional planning aims to simultaneously solve sub-tasks of action planning and topic planning, which plans a reasonable dialogue path consisting of a sequence of dialogue actions and topics with proper orderings. At each turn, the planned path drives the system to distill necessary knowledge from the grounding domain knowledge and meanwhile guides the system to generate a proper utterance. We describe the details of the target-constrained bidirectional planning in Section \ref{sec:trip} and plan-guided dialogue generation in Section \ref{sec:pedg}, respectively.

\begin{figure*}[th!]
\centering
\includegraphics[width=0.98\textwidth]{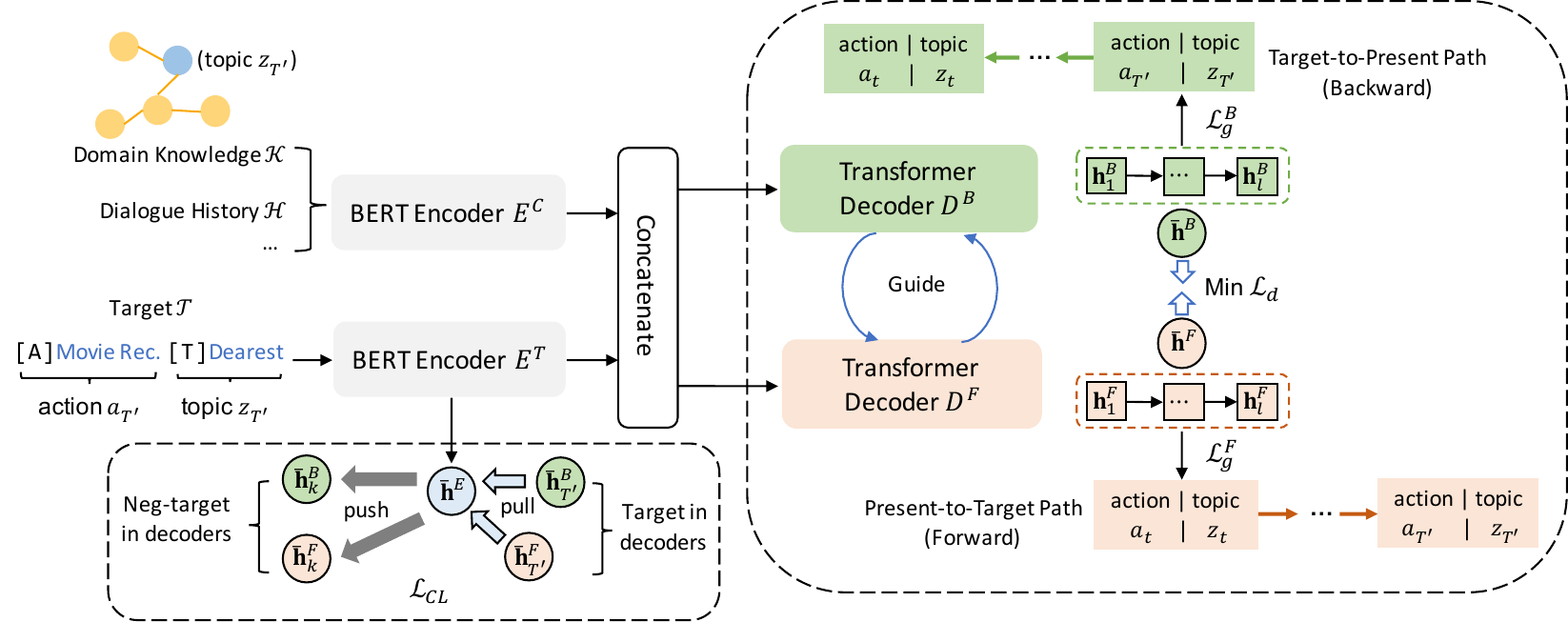}
\caption{Overview of our target-constrained bidirectional planning (TRIP).}
\label{fig:model_plan}
\end{figure*}

\section{Target-constrained Bidirectional Planning}
\label{sec:trip}
In this section, we propose a \textbf{T}a\textbf{R}get-constrained b\textbf{I}directional \textbf{P}lanning (\textbf{TRIP}) model to facilitate the system lead the conversation to achieve the pre-determined target. The overview of our TRIP is shown in Figure \ref{fig:model_plan}. Our TRIP is built with an encoder-decoder architecture, where we adopt two encoders to represent complex input texts and two individual decoders (i.e., a backward decoder and a forward decoder) to complete bidirectional planning.

\subsection{Input Encoding}
\label{sec:encoding}
To efficiently represent various types of input, we take the widely-used pre-trained language model BERT \cite{devlin-etal-2019-bert} as our basic encoder. As shown in Figure \ref{fig:model_plan}, we concatenate the domain knowledge $\mathcal{K}$ and dialogue history $\mathcal{H}$ (and the user profile, if any) as the context. We separate them with a special token \texttt{[SEP]}, which is consistent with the processing in BERT. Then, the context sequence is encoded using a BERT encoder, denoted as $E^{C}$. For the given target consisting of a target action $a_{T^{'}}$ and a target topic $z_{T^{'}}$, we refer to the concatenated text of $a_{T^{'}}$ and  $z_{T^{'}}$ as the target $\mathcal{T}$. We adopt two new special tokens \texttt{[A]} and \texttt{[T]} to differentiate $a_{T^{'}}$ and  $z_{T^{'}}$, e.g., ``\texttt{[A]} \textit{Movie Recommendation} \texttt{[T]} \textit{Dearest}''. Then, the target $\mathcal{T}$ is encoded using another BERT encoder $E^{T}$. Briefly, the encoding of input is formulated as follows:
\begin{align}
    \mathbf{H}^{C}&=\text{BERT Encoder}~ E^{C}([\mathcal{K};\mathcal{H}]) \\
    \mathbf{H}^{T}&=\text{BERT Encoder}~ E^{T}(\mathcal{T})
\end{align}
where $\mathbf{H}^{C}=(\mathbf{h}_{1},\mathbf{h}_{2},\cdots,\mathbf{h}_{L})\in\mathbb{R}^{d\times{L}}$, $\mathbf{H}^{T}=(\mathbf{h}_{1},\mathbf{h}_{2},\cdots,\mathbf{h}_{L^{'}})\in\mathbb{R}^{d\times{L^{'}}}$, $L$ and $L^{'}$ denote context length and target length respectively, $d$ is the hidden size. Here, both $\mathbf{H}^{C}$ and $\mathbf{H}^{T}$ are token-level hidden representations. To maintain full input information for the subsequent planning, we concatenate $\mathbf{H}^{C}$ and $\mathbf{H}^{T}$ as the final input representation, denoted as $\mathbf{M}=[\mathbf{H}^{C};\mathbf{H}^{T}]$.

\subsection{Backward-Forward Path Generation}
\label{sec:pathgen}
Our TRIP aims to plan a reasonable dialogue path consisting of a set of dialogue actions and topics with proper orderings, and this planning process performs in a generation-based manner. We let TRIP generate a forward (present-to-target) path and a backward (target-to-present) path, respectively. It should be noted that the target action $a_{T^{'}}$ and target topic $z_{T^{'}}$ are bounded at the end of the path to be planned. For example, at $t$-th turn, a forward dialogue path is ``$a_{t}|z_{t}\rightarrow a_{t+1}|z_{t+1}\rightarrow \cdots \rightarrow a_{T^{'}}|z_{T^{'}}$'' while its backward dialogue path is ``$a_{T^{'}}|z_{T^{'}}\rightarrow a_{T^{'}-1}|z_{T^{'}-1}\rightarrow \cdots \rightarrow a_{t}|z_{t}$''. Planning a dialogue path from two opposite directions provides supervision to each other during training, and is expected to derive more reasonable dialogue action-topic pairs that compose the ultimate dialogue path, imitating humankind's bidirectional thinking.

In detail, our TRIP generates the two paths based on the Transformer \cite{vaswani2017attention} decoder architecture. We adopt two individual Transformer decoders $D^{B}$ and $D^{F}$ to generate the backward and forward paths, respectively. Both two decoders take the encoded hidden representation $\mathbf{M}$ as input and then output a dialogue path token by token, i.e., ``\texttt{[A]$a_{1}a_{2}\cdots$[T]$t_{1}t_{2}\cdots$[EOS]}'', in an autoregressive manner. Here, $a_{i}$ denotes an action token, $t_{i}$ denotes a topic token, \texttt{[A]} and \texttt{[T]} are two special tokens shared with the encoder $E^{T}$ to differentiate an action and a topic, \texttt{[EOS]} denotes the end of the path sequence.
For the backward decoder $D^{B}$, suppose the output dialogue path $\mathbf{y}$ is represented in token level, i.e., $\mathbf{y}=(y_{1},y_{2},\cdots,y_{T})$ with a sequence length of $T$, and it is conditioned on the input text sequence (denoted as $\mathbf{x}$), the conditional distribution is approximated as follows:
\begin{align}
   p_{\theta}(y_t|\mathbf{y}_{<t},\mathbf{x})&=\text{softmax}(\mathbf{W}\mathbf{h}_{t}^{B}+\mathbf{b}) \\
   \mathbf{h}_{t}^{B}&=D^{B}(y_{t-1},\mathbf{M})
\end{align}
where $\mathbf{W}\in\mathbb{R}^{d\times{d}}$, $\mathbf{b}\in\mathbb{R}^{d}$ denote trainable parameters. We train the backward decoder $D^{B}$ by minimizing the negative log likelihood for given $N$ observations $\{(\mathbf{x}^{(i)},\mathbf{y}^{(i)})\}_{i=1}^{N}$ as follows:
\begin{equation}
    \mathcal{L}_{g}^{B}(\theta)= -\sum_{i=1}^{N}p(\mathbf{y}^{(i)})\log p_{\theta}(\hat{\mathbf{y}}^{(i)}|\mathbf{x}^{(i)})
\end{equation}
where $p(\mathbf{y}^{(i)})$ is the distribution of the ground-truth path sequence, while $p_{\theta}(\hat{\mathbf{y}}^{(i)})$ is the distribution of the approximated output path sequence, $\theta$ denotes all trainable parameters. Similarly, we train the forward decoder $D^{F}$ following the above equations, with the loss function denoted as $\mathcal{L}_{g}^{F}$.

\paragraph{\textbf{Reducing Gap between Backward-Forward Paths}}
Although the backward and forward paths are different, agreement on the dialogue actions and topics derived from the two paths is necessary since the two paths are planned for the same dialogue. By minimizing the decision gap between the backward path and the forward path, the two decoders (i.e., $D^{B}$ and $D^{F}$) are expected to provide supervision to each other and converge on consistent dialogue actions and topics. In detail, we adopt the composition of a linear transformation with the ReLU \cite{nair2010rectified} activation function and an average pooling to obtain the fixed-sized representation of a path, given by:
\begin{align}
    \label{eq:proj}
   \bar{\mathbf{h}}^{B}&=f(\mathbf{V}^{B}),~ \bar{\mathbf{h}}^{F}=f(\mathbf{V}^{F}) \\
   \label{eq:trans}
   f(\mathbf{V})&=\text{AvgPool}([\mathbf{v}_{1}\cdots\mathbf{v}_{T}]), \text{where}~\mathbf{v}_{t}=\text{ReLU}(\mathbf{W}_{1}\mathbf{h}_{t}+\mathbf{b}_{1})
\end{align}
where $\mathbf{W}_{1}\in\mathbb{R}^{d\times{d}}$, $\mathbf{b}_{1}\in\mathbb{R}^{d}$ denote trainable parameters, $\mathbf{h}_{t}$ stands for decoder hidden state. $\bar{\mathbf{h}}^{B}\in\mathbb{R}^{d}$ and $\bar{\mathbf{h}}^{F}\in\mathbb{R}^{d}$ are fixed-sized representation of the backward path and the forward path, respectively. Then, we reduce the gap between the two paths by minimizing $L_{2}$ distance between $\bar{\mathbf{h}}^{B}$ and $\bar{\mathbf{h}}^{F}$ as follows:
\begin{equation}
    \mathcal{L}_{d}=||\bar{\mathbf{h}}^{B}-\bar{\mathbf{h}}^{F}||_{2}
\label{eq:ld}
\end{equation}
where the distance $\mathcal{L}_{d}$ is added to the training loss as a regularization term.

\paragraph{\textbf{Contrastive Generation of Targets}}
\label{para:contgen}
Since our dialogue path generation model is trained with teacher forcing and never exposed to incorrectly generated actions or topics during training, it is insufficient to distinguish between the given target action/topic and other actions or actions. Hence, the model may struggle to constrain the given target generated in the path. To remedy such a situation, we propose a contrastive generation framework (see Figure \ref{fig:model_plan}) to expose the model to various incorrect output targets for a given input target $\mathcal{T}$. Following the contrastive learning framework \cite{lee2021contrastive} for conditional text generation, we train the model to learn the representations of the ground-truth dialogue path by contrasting the positives with the negatives. The critical difference is that, we construct the perturbed negative examples by replacing the target topic in the ground-truth path with multiple randomly sampled topics $\{z_{k}\}_{k=1}^{K}$ $(z_{k}\neq z_{T^{'}})$ from the domain knowledge $\mathcal{K}$, such that the training paths are difficult for the model to distinguish correctly. By identifying which features make the output path negative, these perturbed negative examples are expected to leverage encoders and decoders to learn an adequate representation of the target. It tries to enable our model to generate the necessary target in the path accordingly.

In detail, for the text span consisting of the target action $a_{T^{'}}$ and topic $z_{T^{'}}$ separated with the special tokens \texttt{[A]} and \texttt{[T]} in the two decoders, we project their hidden representations into the latent space following Eq. (\ref{eq:proj}) and Eq. (\ref{eq:trans}), obtaining fixed-sized target representations $\bar{\mathbf{h}}_{T^{'}}^{B}$ and $\bar{\mathbf{h}}_{T^{'}}^{F}$ respectively. Similarly, for the constructed negative examples, we also project those negative targets into the latent space following Eq. (\ref{eq:proj}) and Eq. (\ref{eq:trans}), obtaining corresponding neg-target representations. Since the pre-determined target $\mathcal{T}$ is encoded by the encoder $E^{T}$, we pull the ground-truth target representations in the decoders to the encoded target representation while pushing the neg-target representations in the decoders far away from the encoded target representation (see Figure \ref{fig:model_plan}). Then, we maximize the similarity between the pair of the encoder-decoder targets, while minimizing the similarity between the negative pairs as follows:
\begin{align}
    \mathcal{L}_{CL}^{B}&= -\log{\frac{\text{exp}(\text{sim}(\bar{\mathbf{h}}^{E},\bar{\mathbf{h}}_{T^{'}}^{B})/\tau)}{\sum_{\bar{\mathbf{h}}_{k}^{B}\in{S^{B}}}\text{exp}(\text{sim}(\bar{\mathbf{h}}^{E},\bar{\mathbf{h}}_{k}^{B})/\tau)}} \\
     \mathcal{L}_{CL}^{F}&= -\log{\frac{\text{exp}(\text{sim}(\bar{\mathbf{h}}^{E},\bar{\mathbf{h}}_{T^{'}}^{F})/\tau)}{\sum_{\bar{\mathbf{h}}_{k}^{F}\in{S^{F}}}\text{exp}(\text{sim}(\bar{\mathbf{h}}^{E},\bar{\mathbf{h}}_{k}^{F})/\tau)}}
\end{align}
where $\bar{\mathbf{h}}^{E}$ denotes the averaged representation of the target $\mathcal{T}$ in the encoder $E^{T}$ after transformation following Eq. (\ref{eq:proj}) and Eq. (\ref{eq:trans}), $\bar{\mathbf{h}}_{T^{'}}^{B}$ and $\bar{\mathbf{h}}_{T^{'}}^{F}$ are the ground-truth target representations in the two decoders, respectively. $S^{B}$ and $S^{F}$ stand for a set of neg-target representations in the two decoders, respectively. $\text{sim}(\cdot,\cdot)$ is a cosine similarity function, $\tau$ is a temperature coefficient. Furthermore, we use the averaged result between $\mathcal{L}_{CL}^{B}$ and $\mathcal{L}_{CL}^{F}$ as the contrastive generation loss:
\begin{equation}
    \mathcal{L}_{CL}=\frac{1}{2}(\mathcal{L}_{CL}^{B}+\mathcal{L}_{CL}^{F})
\end{equation}

\paragraph{\textbf{Training}}
During training, we train our TRIP model by minimizing all the losses introduced above. We use two hyperparameters $\beta$ and $\gamma$ to control the importance of gap reducing and contrastive generation, given by:
\begin{equation}
    \mathcal{L}=\mathcal{L}_{g}^{B}+\mathcal{L}_{g}^{F}+\beta\mathcal{L}_{d}+\gamma\mathcal{L}_{CL}
\end{equation}

\subsection{Target-constrained Decoding}
\label{sec:decoding}
After training is done, our TRIP model can be directly used to generate a dialogue path consisting of a set of dialogue actions and topics during inference. Alternatively, we can either use the forward decoder $D^{F}$ to generate a path from the present to the target (denoted as ``forward generation''), or use the backward decoder $D^{B}$ to generate a path from the target to the present (denoted as ``backward generation''). In order to take advantage of the bidirectional decoders, we propose a simple yet effective target-constrained decoding algorithm with a bidirectional agreement based on the widely-used beam search decoding algorithm.

First, each dialogue path is desired to be generated with lexical constraints, i.e., the target action and the target topic should be generated at the end of the path for ``forward generation'' while generated at the beginning of the path for ``backward generation''. To this end, we adopt two additional strategies to fulfill the lexical constraints. For the forward decoder $D^{F}$, we employ the dynamic beam allocation (DBA) \cite{post2018fast} algorithm with a beam size of $k$ to perform lexically constrained decoding, where the required constraint is defined as the given target action and topic. For the backward decoder $D^{B}$, we directly take the target tokens (i.e., a text span consisting of the target action and topic separated with the special tokens \texttt{[A]} and \texttt{[T]}) as the beginning input of the decoder, and then employ vanilla beam search decoding with the same beam size of $k$.

\begin{figure*}[t!]
\centering
\includegraphics[width=0.74\textwidth]{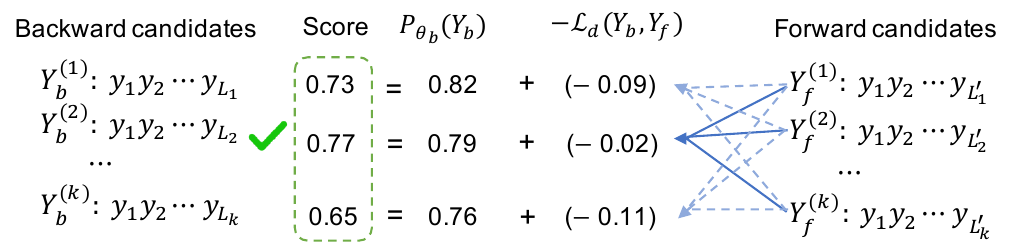}
\caption{Illustration of our target-constrained beam search decoding with bidirectional agreement.}
\label{fig:model_bibs}
\end{figure*}

Second, after the two decoders finish the search process, we obtain $k$ backward candidates (i.e., path sequences) and $k$ forward candidates. As shown in Figure \ref{fig:model_bibs}, to select the best one path sequence as the decoding output, we rank the backward candidates by the following scoring function:
\begin{equation}
    Y_{b}=\mathop{\arg\max}_{Y_{b}^{(i)}\in{S}}P_{\theta_{b}}(Y_{b}^{(i)})+\lambda\cdot(-\frac{1}{k}\sum_{j=1}^{k}\mathcal{L}_{d}(\bar{\mathbf{h}}(Y_{b}^{(i)}),\bar{\mathbf{h}}(Y_{f}^{(j)})))
\label{eq:bibs}
\end{equation}
where $S$ denotes a set of backward candidates, $P_{\theta_{b}}(Y_{b}^{(i)})$ denotes the likelihood of the candidate $Y_{b}^{(i)}$. $\mathcal{L}_{d}(\cdot,\cdot)$ denotes $L_{2}$ distance between a backward candidate and a forward candidate, which is obtained by passing each pair of backward-forward candidates into the model and computed following Eq. (\ref{eq:ld}). Intuitively, the above scoring function ranks the backward candidates by likelihood and gives a partial reward to candidates that satisfy higher agreement (i.e., shorter distance) with the forward candidates, which reduces the gap between inference and training and facilitates the model to select a better one. Here, $\lambda$ is a hyperparameter controlling the weight of the reward term.
Note that we can also select the best one path sequence from the forward candidates using a similar scoring function, which performs a little inferior in most cases in our preliminary experiments. Therefore, by default, we select the best dialogue path sequence from the backward candidates as the ultimate planning output using Eq. (\ref{eq:bibs}).

\section{Plan-guided Dialogue Generation}
\label{sec:pedg}
As mentioned in the preliminaries, we adopt the planned dialogue path (denoted as ``plan path'' $\mathcal{P}$ for short) to guide dialogue generation in a pipeline manner. We expect these plan paths can help a dialogue system distill necessary knowledge and steer the system to generate more proper utterances with control. To achieve plan-guided dialogue generation, we devise two variants and describe them below.

\subsection{Prompt-based Generation}
\label{sec:promptgen}
\begin{figure*}[t!]
\centering
\includegraphics[width=0.46\textwidth]{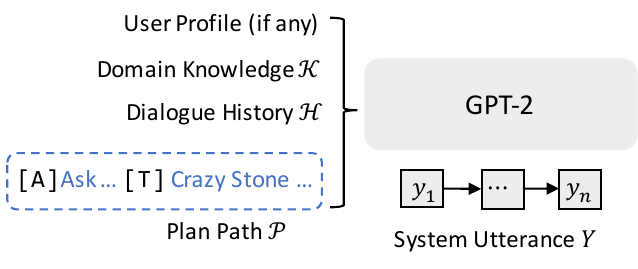}
\caption{Overview of our prompt-based dialogue generation.}
\label{fig:model_dial_a}
\end{figure*}

Motivated by previous works that employ prompt-based learning for dialogue systems \cite{zheng2021exploring,madotto2021few}, we regard each plan path $\mathcal{P}$ as the natural language prompt and then adopt a pre-trained generative language model (LM), e.g., GPT-2 \cite{radford2019language}, for dialogue generation. The overview of our prompt-based dialogue generation is shown in Figure \ref{fig:model_dial_a}. Formally, the plan path $\mathcal{P}$ is concatenated into the given dialogue history $\mathcal{H}$ and domain knowledge $\mathcal{K}$ (and the user profile, if any), formulating the input context $X$ as follows:
\begin{equation}
  X=[\mathcal{K};\mathcal{H};\mathcal{P}]
\label{eq:context}
\end{equation}
where ``$;$'' denotes concatenation. Here, the plan path $\mathcal{P}$ provides essential information that outlines how to achieve the target step by step. With the power of pre-trained LMs, the plan path $\mathcal{P}$ aims to distill necessary knowledge from both input context and LMs. In particular, the input context $X$ is fed into the pre-trained GPT-2 \cite{radford2019language} model to generate the system utterance $Y=\{y_{t}\}_{t=1}^{n}$, where $y_{t}$ is given by:
\begin{equation}
    y_{t}=\text{GPT-2}(y_{<t}, X)
\end{equation}
We fine-tune GPT-2 for a few epochs using ground-truth plan paths in the dataset during training, while we adopt the plan paths generated by our TRIP model during inference.

\subsection{Plan-controlled Generation}
\label{sec:controlgen}
\begin{figure*}[th!]
\centering
\includegraphics[width=0.66\textwidth]{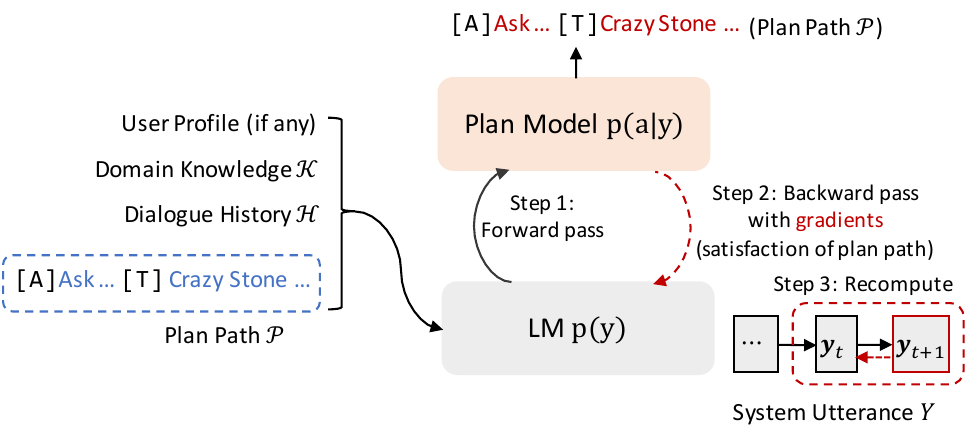}
\caption{Overview of our plan-controlled dialogue generation.}
\label{fig:model_dial_b}
\end{figure*}

Considering that the plan path $\mathcal{P}$ outlines how to achieve the target step by step with a sequence of dialogue actions and topics, we expect to better leverage such critical information to control the attribute (e.g., switching or target topics) of dialogue generation. Inspired by plug-and-play language models for controllable language generation \cite{dathathri2020plug}, we propose a plan-controlled dialogue generation method (see Figure \ref{fig:model_dial_b}). Built upon the pre-trained LM $p(y)$, e.g., GPT-2, we employ a simple plan model $p(a|y)$ to act as the attribute controller, which guides the generation of the LM $p(y)$ through gradients. Considering that the generation of system utterances follows the conditional form of $p(y|a)\propto p(y)\cdot p(a|y)$, we shift the hidden states of generation in the direction of the sum of two gradients: one toward higher log-likelihood of the unmodified LM $p(y)$ and one toward higher log-likelihood of the attribute $a$ under the conditional plan model $p(a|y)$. Combining the two factors together makes it controllable to guide dialogue generation in a given direction (i.e., the plan path $\mathcal{P}$) with specified strength.

Concretely, as shown in Figure \ref{fig:model_dial_b}, we take the concatenated context $X$ following Eq. (\ref{eq:context}) as input and employ the pre-trained GPT-2 (denoted as LM $p(y)$) for dialogue generation. Let us define the cached hidden representations $\mathbf{H}_{t}$ of the LM $p(y)$ as all key-value pairs from the past, i.e., $\mathbf{H}_{t}=[(\mathbf{K}_{t}^{(1)},\mathbf{V}_{t}^{(1)}),\cdots,(\mathbf{K}_{t}^{(l)},\mathbf{V}_{t}^{(l)})]$, where $(\mathbf{K}_{t}^{(i)},\mathbf{V}_{t}^{(i)})$ corresponds to the key-value pairs from the $i$-th layer generated at all time-steps from 0 to $t$. Efficient computations of the LM $p(y)$ to generate the next token $y_{t+1}$ using the cached $\mathbf{H}_{t}$ are summarized as:
\begin{align}
    \mathbf{o}_{t+1},\mathbf{H}_{t+1}&=\text{LM}(y_{t},\mathbf{H}_{t}) \\
    \label{eq:gen}
    y_{t+1}\sim p(y_{t+1})&=\text{softmax}(\mathbf{W}\mathbf{o}_{t+1})
\end{align}
where $\mathbf{W}$ is a linear transformation that maps the hidden vector $\mathbf{o}_{t+1}$ to a vector of vocabulary size. On top of that, we build a simple plan model (denoted as $p(a|y)$) using a Transformer \cite{vaswani2017attention} decoder. The plan model $p(a|y)$ aims to re-generate the given plan path $\mathcal{P}$ conditioning on hidden vectors $\{\mathbf{o}_{0},\mathbf{o}_{1},\cdots,\mathbf{o}_{t}\}$ of the LM $p(y)$ across all time-steps from 0 to $t$. Here, the plan model $p(a|y)$ performs as a generative discriminator that gives the LM $p(y)$ a higher reward for having the desired generation direction, i.e., the plan path $\mathcal{P}$.
During training, we jointly train the plan model $p(a|y)$ and fine-tune the LM $p(y)$ by maximizing log-likelihood.

During inference, we use the plan model $p(a|y)$ to control the output of the LM $p(y)$ at every generation step $t$, following \cite{dathathri2020plug}. As shown in Figure \ref{fig:model_dial_b}, a forward pass is performed first through the LM $p(y)$ to compute the unmodified likelihood. In step 2, a backward pass updates the cached key-value pairs $\mathbf{H}_{t}$ with gradients from the plan model $p(a|y)$. Let $\Delta\mathbf{H}_{t}$ be the update to $\mathbf{H}_{t}$, such that the generation with $(\mathbf{H}_{t}+\Delta\mathbf{H}_{t})$ shifts the distribution of the generated utterance so that it is more likely to satisfy the plan path $\mathcal{P}$. $\Delta\mathbf{H}_{t}$ is initialized at zero and updated as follows:
\begin{equation}
    \Delta\mathbf{H}_{t} \leftarrow \Delta\mathbf{H}_{t}+\alpha\frac{\nabla_{\Delta\mathbf{H}_{t}}\log{p(a|\mathbf{H}_{t}+\Delta\mathbf{H}_{t})}}{\lVert\nabla_{\Delta\mathbf{H}_{t}}\log{p(a|\mathbf{H}_{t}+\Delta\mathbf{H}_{t})}\rVert}
\end{equation}
where $\alpha$ is the step size. This updating step can be repeated multiple times while in practice we update once for computational efficiency. Subsequently, we use the updated key-value pairs to recompute the perturbed hidden vector $\widetilde{\mathbf{o}}_{t+1}$, given by:
\begin{align}
    \widetilde{\mathbf{o}}_{t+1},\mathbf{H}_{t+1}&=\text{LM}(y_{t},\widetilde{\mathbf{H}}_{t}), \text{where}~\widetilde{\mathbf{H}}_{t}=\mathbf{H}_{t}+\Delta\mathbf{H}_{t}
\end{align}
The perturbed $\widetilde{\mathbf{o}}_{t+1}$ is then used to generate the next token $y_{t+1}$ following Eq. (\ref{eq:gen}).

\section{Experimental Setup}

\subsection{Datasets and Processing}

\paragraph{\textbf{Datasets}}
The task of target-oriented dialogue generation is still relatively under-explored. Although many publicly available dialogue datasets exist, we find the DuRecDial \cite{liu-etal-2020-towards-conversational} and DuRecDial 2.0 \cite{liu2021durecdial} are the most suitable datasets for this task to the best of our knowledge. The system often leads the dialogue proactively instead of passively responding to users in the DuRecDial and DuRecDial 2.0 datasets, with rich interactive actions such as chitchat, question answering, recommendation, etc. We first briefly introduce the two datasets and then introduce how we re-purpose the datasets for target-oriented dialogue generation.

The original DuRecDial and DuRecDial 2.0 datasets were collected from crowdsourced human-to-human dialogues. One person was defined as the seeker (the user's role) and the other as the recommender (the system's role) in a dialogue. The recommender was required to proactively lead the dialogue and make recommendations by introducing new topics. Each seeker was equipped with a user profile containing user attributes (e.g., name, age range) and his/her past preference information. In order to perform smooth conversations with the seeker, the recommender has a domain knowledge graph consisting of domain-specific topics (e.g., movies, music, and food) with related attributes. More importantly, a dialogue path composed of dialogue actions and topics was annotated for the recommender (or the system) from the beginning to the end of the dialogue. The original DuRecDial dataset contains about 10k multi-turn Chinese dialogues and 156k utterances, while the DuRecDial 2.0 dataset has 8.2k dialogues aligned across English and Chinese languages. In this work, we adopt the DuRecDial dataset in Chinese and the DuRecDial 2.0 dataset in English for experiments.

\begin{figure*}[t!]
\centering
\subfigure[]{
    \begin{minipage}[b]{0.48\textwidth}
	    \includegraphics[width=1\linewidth]{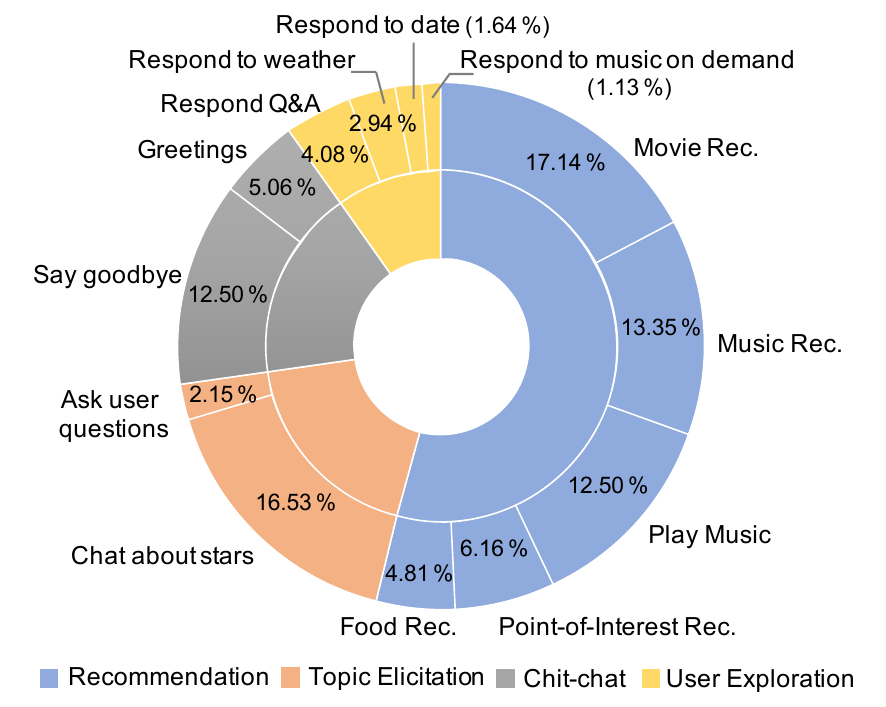}
    \end{minipage}
    \label{fig:stats.a}
}
\subfigure[]{
    \begin{minipage}[b]{0.48\textwidth}
	    \includegraphics[width=1\linewidth]{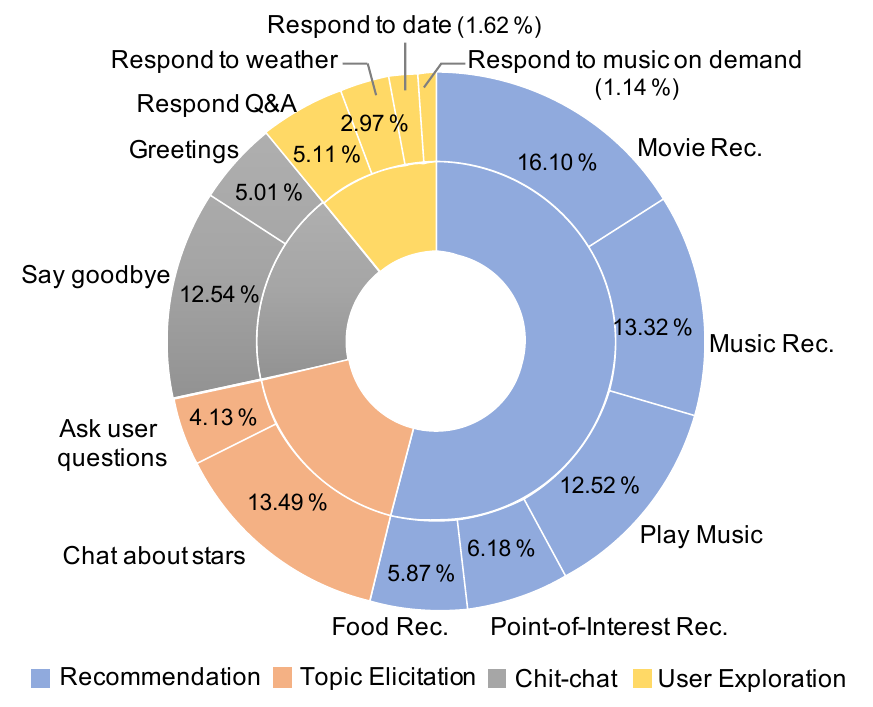}
    \end{minipage}
    \label{fig:stats.b}
}
\caption{Statistics of the system's dialogue actions on the re-purposed (a) DuRecDial and (b) DuRecDial 2.0 datasets.}
\label{fig:action_stats}
\end{figure*}

\begin{table}[t!]
\caption{Statistics of the re-purposed DuRecDial and DuRecDial 2.0 datasets. Here, ``Dial.'' denotes ``dialogue'', ``Utter.'' denotes ``utterance'', ``ID'' and ``OOD'' are short for ``In-Domain'' and ``Out-Of-Domain'', respectively.}
\centering
\resizebox{0.86\linewidth}{!}{
\begin{tabular}{ll|c|c|c|c|c|c|c}
\toprule
 \multicolumn{2}{c|}{\multirowcell{2}{\textbf{Dataset}}}  & \multirow{2}{*}{\textbf{\#Dial.}} & \multirow{2}{*}{\textbf{\#Utter.}} & \multirowcell{2}{\textbf{Plan Path} \\ \textbf{\#Avg.}} & \multicolumn{2}{c|}{\textbf{Dial. Turn}} &  \multicolumn{2}{c}{\textbf{Knowledge Triples}} \\
 &  &  &   &   &    \textbf{\#Max.} &  \textbf{\#Avg.}  & \textbf{\#Max.} & \textbf{\#Avg.} \\
\midrule
 \multirow{4}{*}{DuRecDial} & Train  & 4,440  & 72,466  &  4.4 & 13 & 8.2  & 47  & 35.1 \\
 & Dev   &  633 &  10,467  &  4.5 & 12  & 8.3  & 46 &  35.2 \\
 & Test-ID  &  780 &  12,633 & 4.4  & 13  & 8.1 & 45  & 34.8 \\
 & Test-OOD  &  486 &  7,500 & 4.3  & 14 & 7.7 & 43 & 34.1 \\
\midrule
 \multirow{4}{*}{DuRecDial 2.0} & Train  & 4,256  & 68,781  &  4.4 & 13  & 8.1  & 69  & 55.0 \\
 & Dev   &  608 &  9,677  &  4.3 & 14  & 8.0  & 68 & 55.4 \\
 & Test-ID  &  770 &  12,299 & 4.3  & 13  & 8.0 & 69 & 55.0  \\
 & Test-OOD  &  446 &  7,962 & 4.8  & 12  & 8.9 & 68  & 56.0  \\
\bottomrule
\end{tabular}}
\label{tab:data_stats}
\end{table}

\paragraph{\textbf{Data Processing}}
Since no explicit targets are annotated in the original DuRecDial and DuRecDial 2.0 datasets, we re-purpose the two datasets through automatic target construction for target-oriented dialogue generation, following \cite{wang2022follow}. For all those dialogues that are proactively led by the system, we treat the topic that the user has accepted at the end of each dialogue as the target topic, and view the system's corresponding action (e.g., movie recommendation, point-of-interest recommendation, etc.) as the target action. We filter out those dialogues without introducing any new recommendation topic. In addition, we discard all user reviews in the original domain knowledge triples because user reviews do not belong to domain knowledge. We further enrich existing grounding domain knowledge triples corresponding to each dialogue with more knowledge triples sampled from the triples within two hops of the target topics in the dataset. Hence, it is more challenging for knowledge selection and topic planning. Note that each target topic is guaranteed to ground on the domain knowledge triples corresponding to the dialogue. Statistics of all the system's dialogue actions on the re-purposed DuRecDial and DuRecDial 2.0 datasets are shown in Figure \ref{fig:action_stats}. The total numbers of topics are 640 (including a \texttt{NULL} topic) and 628 (including a \texttt{NULL} topic) in the DuRecDial and DuRecDial 2.0, respectively.

Following the splitting criterion in \cite{liu-etal-2020-towards-conversational,wang2022follow}, we split the re-purposed DuRecDial dataset into the train/dev/test sets with 4,440/633/1,266 dialogues, respectively. Similarly, we obtain the train/dev/test sets of the DuRecDial 2.0 dataset with 4,256/608/1,216 dialogues, respectively. To investigate the performance of different methods for target-oriented dialogue generation, we further use the processed datasets with two types of splits for the test set: 1) \textbf{In-Domain (ID)} split and 2) \textbf{Out-Of-Domain (OOD)} split, similar to \cite{sevegnani-etal-2021-otters,gupta-etal-2022-target}. The OOD split ensures that none of the target topics in the test set are present in the train set. In the ID split, the target topics in the test set are allowed to appear in the train set. In total, statistics of the two re-purposed datasets are reported in Table \ref{tab:data_stats}. We can observe an average of 4.3 $\sim$ 4.8 action-topic transitions (i.e., the average length of the plan path) from the beginning toward the target.

\subsection{Baseline Methods}
\label{baselines}

\paragraph{\textbf{Dialogue Generation}}
To validate the effectiveness of our proposed two variants for target-oriented dialogue generation, we first compare them with the following dialogue generation methods based on pre-trained language models:
\begin{itemize}
\item \textit{DialoGPT} \cite{zhang-etal-2020-dialogpt}: It is an autoregressive generation model pre-trained using large-scale dialogue corpora for conversational response generation. We adopt the pre-trained model\footnote{https://huggingface.co/microsoft/DialoGPT-small} for fine-tuning the dataset in English. For fine-tuning the dataset in Chinese, we adopt the Chinese version \cite{wang2020large} pre-trained model\footnote{https://github.com/thu-coai/CDial-GPT}.
\item \textit{GPT-2} \cite{radford2019language}: It is a pre-trained autoregressive generation model for language generation. We use the publicly available GPT-2 base\footnote{https://huggingface.co/gpt2} model and Chinese GPT-2 base\footnote{https://huggingface.co/uer/gpt2-chinese-cluecorpussmall} model for fine-tuning the English and Chinese datasets, respectively.
\item \textit{BART} \cite{lewis2020bart}: It is an encoder-decoder pre-trained language model with denoising for natural language generation. We use the publicly available BART-base\footnote{https://huggingface.co/facebook/bart-base} model and Chinese BART-base\footnote{https://huggingface.co/fnlp/bart-base-chinese} model for fine-tuning the English and Chinese datasets, respectively.
\end{itemize}
Note that these models concatenate all parts of input texts described in the problem definition as the model input and are fine-tuned to generate system utterances directly.

We also compare our plan-guided dialogue generation methods with several competitive models that are closely related to target-oriented dialogue generation, where they follow the predict-then-generate paradigm or planning-enhanced generation paradigm:
\begin{itemize}
\item \textit{MGCG\_G} \cite{liu-etal-2020-towards-conversational}: It employs the predicted next dialogue action and next topic to guide system utterance generation. Following our problem setting, we re-run the officially released code\footnote{https://github.com/PaddlePaddle/Research/tree/master/NLP/ACL2020-DuRecDial} on the two re-purposed datasets.
\item \textit{KERS} \cite{zhang-etal-2021-kers}: It has a knowledge-enhanced mechanism for recommendation dialogue generation built upon Transformer \cite{vaswani2017attention} architectures. Similarly, we re-run the officially released code\footnote{https://github.com/z562/KERS} on the two re-purposed datasets.
\item \textit{TCP-Dial} \cite{wang2022follow}: It proposes a target-driven conversation planning method to explicitly extract necessary knowledge and then guides dialogue generation built upon various backbone models. We adopt the GPT-2 \cite{radford2019language} as the backbone model for comparisons in this work, and re-run the officially released code\footnote{https://github.com/iwangjian/Plan4RecDial} on the two re-purposed datasets.
\end{itemize}

\paragraph{\textbf{Dialogue Planning}}
To explore the performance of planning for target-oriented dialogue systems, we compare our TRIP model with the following dialogue planning methods: 
\begin{itemize}
\item \textit{MGCG} \cite{liu-etal-2020-towards-conversational}: It employs a convolutional neural network \cite{kim-2014-convolutional} to conduct multi-task predictions for the next dialogue action and the next topic. However, it assumes that ground-truth historical dialogue actions and topics are known for a system. In this work, we only provide the target (i.e., a target action paired with a target topic), while the system itself should plan all interim dialogue actions and topics to achieve the target. For a fair comparison, we take the same input as our problem definition to conduct multi-task predictions.
\item \textit{KERS} \cite{zhang-etal-2021-kers}: It aims to generate the next dialogue action and the next topic based on a Transformer \cite{vaswani2017attention} network. Similarly, we take the same input as our problem definition for KERS.
\item \textit{BERT} \cite{devlin-etal-2019-bert}: Based on the intuition of multi-task predictions, we fine-tune the widely-used pre-trained language model BERT \cite{devlin-etal-2019-bert} by adding two fully-connected layers to jointly predict the system's next dialogue action and topic. We use the publicly available BERT-base-uncased\footnote{https://huggingface.co/bert-base-uncased} model and the Chinese BERT-base\footnote{https://huggingface.co/bert-base-chinese} model for fine-tuning the English and Chinese datasets, respectively.
\item \textit{TCP} \cite{wang2022follow}: It is a target-driven planning framework that aims to plan a path consisting of dialogue actions and topics in a generation-based manner. To the best of our knowledge, TCP is the most related work to ours on dialogue planning for the target-oriented dialogue generation task.
\end{itemize}

\subsection{Evaluation Metrics}

\paragraph{\textbf{Automatic Evaluation}}

Following many previous studies \cite{liu-etal-2020-towards-conversational,wang2022follow} in dialogue generation, we adopt widely-used metrics for automatic evaluation as follows:
\begin{itemize}
    \item \textit{Perplexity} (\textit{PPL}) and \textit{distinct} (\textit{DIST}) \cite{li-etal-2016-diversity}: The \textit{perplexity} and \textit{distinct} measure the fluency and the diversity of generated system utterances, respectively.
    \item \textit{F1}: The \textit{F1} score estimates the precision and recall of each generated utterance at the word level (the character level if evaluating Chinese datasets).
    \item \textit{BLEU} \cite{papineni-etal-2002-bleu}: The \textit{BLEU} score calculates $n$-gram overlaps between generated utterances and gold utterances.
    \item  \textit{Knowledge F1} (\textit{Know. F1}) \cite{liu-etal-2020-towards-conversational}: It evaluates the performance of generating correct knowledge (e.g., topics, attributes) from the domain knowledge triples. However, there is no labeled knowledge annotated in gold system utterances in the datasets. We first conduct strict string matching to search for the entities from the domain knowledge that also occur in each gold system utterance as the knowledge label. Since some knowledge entries (\textit{object} in the triple $\langle$\textit{subject, relation, object}$\rangle$) are in form of long texts (e.g., topic-associated attributes) and they can be paraphrased during conversations, we then compute word-based recall scores between knowledge entries and gold system utterances. We take the knowledge entries whose recall scores are greater than a threshold of 0.55 as the pseudo label. For evaluating knowledge F1, we take the same threshold (i.e., 0.55) to examine whether a knowledge entry is hit in the generated utterances.
    \item \textit{Goal success rate} (\textit{Goal Succ.}): It is essential to validate a model of how well it achieves the pre-determined target, where the target topic can be used for automatic evaluation. Similar to \cite{wang2022follow}, we choose the dialogues at the target turn in the test dataset to compute the ratio of generating the target topic correctly for each model as the \textit{goal success rate}.
\end{itemize}

To evaluate dialogue planning, we adopt the following metrics:
\begin{itemize}
    \item \textit{F1}: It estimates the micro-averaged precision and recall of the predicted action or topic. For generation-based models, we take the generated action or topic at the evaluating turn for a fair comparison. We report dialogue action F1 and topic F1 scores in the experimental results, respectively.
    \item \textit{Bigram F1} (\textit{Bi. F1}): Due to the nature of dialogues, multiple temporary planning strategies can be reasonable before completing the target. Following \cite{zhou2020augmenting}, we also expand gold labels by taking the system's actions and topics within the previous turn and the following turn into account, formulating the \textit{bigram F1}.
\end{itemize}

\paragraph{\textbf{Human Evaluation}}

Similar to \cite{liu-etal-2020-towards-conversational}, we conduct human evaluation from both turn-level and dialogue-level aspects. For turn-level evaluation, we randomly select 50 samples from the test-ID dataset and 50 samples from the test-OOD dataset and ask each model to produce system utterances. Three well-educated annotators are required to mark scores for different models from the aspects of both  \romannumeral1) \textit{appropriateness} and  \romannumeral2) \textit{informativeness}. The \textit{appropriateness} measures if a generated system utterance can complete the current plan and respond to the dialogue context appropriately, and the \textit{informativeness} measures if a model can make full use of the grounding domain knowledge to generate an informative utterance. For fairness, all model names are masked to annotators during the evaluation process.

For dialogue-level evaluation, we let each model interact with human annotators, which indicates that a model's generated utterance in the current turn will be further used as a part of the dialogue history in the next turn. To ensure that the evaluation covers a wide range of targets, we randomly sample 5 different target actions from the test sets, with each action consisting of 10 different target topics. In total, 50 different dialogue targets are evaluated. To examine whether a model can lead the conversation to achieve the pre-determined target proactively and smoothly, we do not expose the target to human annotators during human-model interactions. Besides, human annotators are asked to be consistent with each given user profile, if any. All human-model dialogues are limited to no more than 15 turns. At the end of each dialogue, we will expose the pre-determined target to human annotators and ask each annotator to mark scores for different models from \romannumeral1) \textit{proactivity}, which measures if a model can proactively lead new actions and topics in the conversation, \romannumeral2) \textit{coherence}, which measures the overall fluency and naturalness of the whole dialogue generation, and \romannumeral3) \textit{goal success}, which estimates how well the pre-determined target is achieved.

For all the above metrics, human evaluation scores are settled from \{0, 1, 2\}, where a higher score denotes better performance. The agreement among the annotators is measured by Fleiss's kappa \cite{fleiss1971measuring}. The averaged score of different human annotators is reported as the ultimate human evaluation result for each model.

\subsection{Implementation Details}

Our TRIP model and plan-guided generation methods are implemented by PyTorch. During planning, we adopt the BERT-base model (12 layers, 768 dimensions, 12 heads, and 110M parameters) and the Chinese BERT-base model released in Huggingface's Transformers \cite{wolf-etal-2020-transformers} library as input encoders for the DuRecDial 2.0 dataset and the DuRecDial dataset, respectively. Both the backward and forward decoders are stacked to 6 layers with 8 attention heads. The hidden size is set to be consistent with BERT encoders, i.e., 768. The embeddings of the two decoders are randomly initialized, with the vocabulary size consistent with the BERT encoders. For the contrastive generation of targets, we randomly sample 3 negatives. The temperature coefficient $\tau$ is set to 0.1. The two hyperparameters $\beta$ and $\gamma$ are set to 0.1 and 1.0, respectively. We set the batch size to 6 due to memory constraints and train our TRIP model with a maximum of 10 epochs. We adopt the Adam \cite{kingma2014adam} optimizer with an initial learning rate of $2e\text{-}5$ and warm up over the first 3,000 training steps with linear decay. We select the best model based on the performance of the validation set. For the target-constrained decoding, the beam size is set to 3, with a maximum decoding length of 80. The hyperparameter $\lambda$ that controls the weight of the agreement reward is set to 1.0.

\begin{table*}[t!]
\caption{Evaluation results of dialogue generation on the DuRecDial dataset. Significant improvements over baseline models are marked with * (t-test, $p < 0.05$).}
\centering
\resizebox{0.97\textwidth}{!}{
\begin{tabular}{cl|c|c|c|c|c|c}
\toprule
& \textbf{Model} & \textbf{PPL ($\downarrow$)} & \textbf{F1 (\%)}  & \textbf{BLEU-1 / 2}  & \textbf{DIST-1 / 2}  & \textbf{Know. F1 (\%)} & \textbf{Goal Succ. (\%)} \\
\midrule 
\midrule
\multirow{8}{*}{Test-ID} & MGCG\_G & 17.81  & 40.07 &  0.352 / 0.273  & 0.012 / 0.058  & 41.14  & 37.22  \\
  & KERS  & 12.39 & 38.24  & 0.356 / 0.277  & 0.011 / 0.047  & 45.04  &  44.16 \\
  &  DialoGPT &  5.64 & 34.25  & 0.314 / 0.237  & 0.009 / 0.045  &  33.90  & 38.55 \\
  &  GPT-2 & 4.42 & 39.57  & 0.370 / 0.297 & 0.012 / 0.062 & 45.84  & 59.97  \\
   &  BART &  5.01 & 38.68  & 0.341 / 0.268  & \textbf{0.013} / \textbf{0.075} & 43.84  & 68.31  \\
  & TCP-Dial & 4.41  & 39.10 & 0.379 / 0.303  & 0.011 / 0.058  & 49.90  & 69.88 \\
\cmidrule{2-8}
 & Ours (prompt) & \textbf{4.38} &  42.34* & 0.387* / \textbf{0.312}* & 0.011 / 0.059  & 53.12*  & 77.40*  \\
 & Ours (controlled) & 4.40  &  \textbf{43.11}*  & \textbf{0.388}* / \textbf{0.312}*  & 0.012 / 0.062  & \textbf{53.69}*  & \textbf{77.55}*  \\
\midrule
\midrule
\multirow{8}{*}{Test-OOD} & MGCG\_G & 18.57  &  36.72  & 0.339 / 0.257  &  0.012 / 0.045  &  32.24 & 10.24 \\
  & KERS  &  14.06 &  36.28  & 0.340 / 0.262  &  0.010 / 0.042  &  38.38 &  14.07 \\
  &  DialoGPT & 5.57  & 35.52  & 0.325 / 0.252  & 0.010 / 0.041  &  35.16 & 37.16 \\
  &  GPT-2 & 4.48 &  40.90 &  0.388 / 0.317  & 0.013 / 0.055  &  47.42 &  58.60 \\
  &  BART & 5.04 &  40.04 &  0.360 / 0.288  & \textbf{0.015} / \textbf{0.067}   & 46.62  &  65.45 \\
  & TCP-Dial  & 4.46 & 34.65 & 0.358 / 0.279 & 0.012 / 0.055  & 32.29 & 16.72 \\
\cmidrule{2-8}
 & Ours (prompt) & 4.46 &  41.90*  &  0.396* / \textbf{0.322}*  &  0.012 / 0.054  &  48.32* &  78.30*  \\
 & Ours (controlled) & \textbf{4.45}  & \textbf{42.40}*   & \textbf{0.397}* / \textbf{0.322}*   &  0.013 / 0.055  & \textbf{48.93}*  & \textbf{79.80}*   \\
\bottomrule
\end{tabular}}
\label{tab:dialogue_zh}
\end{table*}

For our plan-guided dialogue generation, we fine-tune the GPT-2 base model and Chinese GPT-2 base model released in Huggingface's Transformers \cite{wolf-etal-2020-transformers} library on the DuRecDial 2.0 dataset and the DuRecDial dataset, respectively. The length of the concatenated input text is limited to 512. In addition, the plan model $p(a|y)$ in the plan-controlled generation employs a lightweight Transformer decoder with 3 layers and 8 attention heads. The embeddings of $p(a|y)$ are copied from the embeddings of the LM $p(y)$ (i.e., GPT-2). The step size $\alpha$ is set to 0.01. Both variants employ greedy search decoding during generation, with a maximum decoding length of 100. All the experiments are conducted on a single NVIDIA GeForce 3090 GPU machine. Our code and data are available at \url{https://github.com/iwangjian/TRIP}.

\section{Experimental Results}
Our experiments and detailed analysis aim to answer the following research questions:
\begin{itemize}[leftmargin=*]
    \item RQ1: How is the performance of the proposed planning for generation on the end task of target-oriented dialogue generation compared to existing methods?
    \item RQ2: How is the performance of the proposed TRIP model on each sub-task, including action planning and topic planning, compared to existing methods?
    \item RQ3: How does each proposed component or strategy contribute to the overall performance?
    \item RQ4: What are the merits and limitations of the pipelined approach in this work? 
\end{itemize}

\begin{table*}[th!]
\caption{Evaluation results of dialogue generation on the DuRecDial 2.0 dataset. Significant improvements over baseline models are marked with * (t-test, $p < 0.05$).}
\centering
\resizebox{0.97\textwidth}{!}{
\begin{tabular}{cl|c|c|c|c|c|c}
\toprule
& \textbf{Model} & \textbf{PPL ($\downarrow$)} & \textbf{F1 (\%)}  & \textbf{BLEU-1 / 2}  & \textbf{DIST-1 / 2}  & \textbf{Know. F1 (\%)} & \textbf{Goal Succ. (\%)} \\
\midrule 
\midrule
\multirow{8}{*}{Test-ID} & MGCG\_G & 25.32  & 35.13 & 0.316 / 0.211  & 0.016 / 0.053  &  39.53 & 20.51  \\
  & KERS  & 20.15  &  31.27 & 0.288 / 0.196 & 0.017 / 0.061  & 41.18  &  28.75 \\
  &  DialoGPT & 5.26  & 35.12 & 0.304 / 0.212 & 0.023 / 0.076  & 42.71  & 30.09 \\
  &  GPT-2 & 5.33 &  36.86 &  0.314 / 0.222 & 0.024 / 0.081 & 43.62 & 31.64 \\
  &  BART &  6.46 & 36.11  & 0.279 / 0.181 & \textbf{0.030} / \textbf{0.096} & 43.33  & 33.05   \\
  & TCP-Dial &  5.88 &  34.46 &  0.293 / 0.201  & 0.027 / 0.091  & 45.75 & 29.49  \\
\cmidrule{2-8}
 & Ours (prompt) & \textbf{5.17} &  37.40* & 0.326* / 0.233* & 0.026 / 0.083  & 47.03*  & 36.13*  \\
 & Ours (controlled) & 5.23  &  \textbf{37.48}*  & \textbf{0.331}* / \textbf{0.238}*  & 0.025 / 0.080  & \textbf{47.44}*  & \textbf{38.67}*  \\
\midrule
\midrule
\multirow{8}{*}{Test-OOD} & MGCG\_G &  28.21 & 30.84 & 0.276 / 0.167  &  0.015 / 0.046  &  20.53 & 5.65 \\
  & KERS  & 24.35  &  27.91  & 0.259 / 0.160  &  0.016 / 0.058  & 26.88  & 11.06  \\
  &  DialoGPT &  5.37 & 31.27  & 0.283 / 0.176  &  0.021 / 0.068 & 30.75  & 26.57 \\
  &  GPT-2 & 5.86 &  31.26 & 0.266 / 0.193  & 0.023 / 0.077  & 28.79  &  26.30 \\
  &  BART & 8.09 & 32.38  & 0.244 / 0.149 & 0.026 / 0.081  & 30.02 & 28.10  \\
  & TCP-Dial  &  8.24 & 29.24  & 0.255 / 0.165 & \textbf{0.027} / \textbf{0.089}  & 21.36 & 6.97 \\
\cmidrule{2-8}
 & Ours (prompt) & 5.63 &  33.05*  &  0.292* / 0.198*  &  0.025 / 0.079  &  31.81* &  31.17*  \\
 & Ours (controlled) & \textbf{5.59}  & \textbf{33.30}*   & \textbf{0.297}* / \textbf{0.202}*   &  0.024 / 0.078  & \textbf{32.82}*  & \textbf{33.44}*   \\
\bottomrule
\end{tabular}}
\label{tab:dialogue_en}
\end{table*}

\subsection{Evaluation Results of Dialogue Generation (RQ1)}
Our automatic evaluation results of dialogue generation on the DuRecDial and DuRecDial 2.0 datasets are reported in Table \ref{tab:dialogue_zh} and Table \ref{tab:dialogue_en}, respectively. The best result in terms of the corresponding metric is highlighted in boldface. As shown in Table \ref{tab:dialogue_zh}, MGCG\_G and KERS are capable of obtaining better results than DialoGPT on the in-domain (ID) test set in terms of F1, BLEU, and DIST. Considering that the two models are trained without using pre-trained language models, their competitive performance mainly benefits from the prediction of the next dialogue action and topic, which guides the model to generate more informative and reasonable utterances. However, MGCG\_G, KERS, and DialoGPT obtain poor goal success rates, which drop sharply on the out-of-domain (OOD) test set in particular. It shows that they still struggle to lead dialogues to reach the target when necessary. In comparison, GPT-2 and BART perform much better than other baseline models over various metrics when evaluated on both in-domain (ID) and out-of-domain (OOD) test sets. We note that in terms of DIST-1/2 scores, BART is significantly better than other baselines because BART seldom generates repeated words, making the generated utterances more diverse in many cases. However, GPT-2 performs better in most cases in generating $n$-gram overlapped utterances (see BLEU-1/2) with correct knowledge (see Know. F1). We employ GPT-2 as our backbone model due to its strong generation ability and ease of incorporation in our plan-controlled generation.
For the TCP-Dial, the goal success rate deteriorates remarkably on the OOD test dataset (16.72\%) compared to the ID test dataset (69.88\%). It is because TCP-Dial explicitly extracts topic-centric knowledge triples according to the planned topic, which may discard necessary domain knowledge when the target topic is not correctly planned especially on the OOD test dataset, making it difficult to generate a proper utterance containing the target topic.

Compared to baseline methods, our proposed plan-guided generation methods achieve significant improvements over most evaluation metrics. For example, our prompt-based generation method achieves much better knowledge F1 scores, i.e., 53.12\% and 48.32\% on the in-domain (ID) and out-of-domain (OOD) test sets (see Table \ref{tab:dialogue_zh}). It shows that our model is more likely to generate correct knowledge (e.g., topics, attributes) from the domain knowledge triples. In terms of the goal success rate according to Table \ref{tab:dialogue_zh}, our prompt-based generation method obtains a much higher score of 77.40\%, which significantly outperforms existing baseline models. It indicates that we successfully stimulate the potential of the existing pre-trained language model (i.e., GPT-2) to generate more proper utterances for target-oriented dialogue generation by enriching appropriate dialogue paths as prompts. More importantly, our model is still able to maintain a high goal success rate when evaluated on the out-of-domain (OOD) test set. In contrast to GPT-2, our model mainly benefits from our dialogue planning, which verifies the effectiveness of the proposed planning for generation on the end task of target-oriented dialogue generation. Moreover, our plan-controlled generation method further improves the performance of the prompt-based generation method, demonstrating that each planned dialogue path can further steer the model by controlling the generation process of each utterance.

\begin{table}[t!]
\caption{Experimental results of dialogue planning on the DuRecDial dataset. Significant improvements over baseline models are marked with * (t-test, $p < 0.05$).}
\centering
\resizebox{0.7\linewidth}{!}{
\begin{tabular}{l|cc|cc|cc|cc}
\toprule
\multirow{3}{*}{\textbf{Model}} &  \multicolumn{4}{c|}{\textbf{Test-ID}} &  \multicolumn{4}{c}{\textbf{Test-OOD}} \\
\cmidrule{2-9}
 & \multicolumn{2}{c|}{\textbf{ Action}} &  \multicolumn{2}{c|}{\textbf{ Topic}} & \multicolumn{2}{c|}{\textbf{ Action}} &  \multicolumn{2}{c}{\textbf{ Topic}}\\
 & \textbf{F1} &  \textbf{Bi. F1}  & \textbf{F1} & \textbf{Bi. F1} & \textbf{F1} &  \textbf{Bi. F1}  & \textbf{F1} & \textbf{Bi. F1} \\
\midrule
MGCG & 87.30 &  93.16 &  71.21  &  76.99 & 88.29 & 91.75  & 44.50  & 49.88  \\
KERS &  92.13 &  94.08  & 80.34  & 82.60 & 90.34 & 91.88  & 41.25  & 44.36 \\
BERT &  94.82 &  95.99  & 83.37  & 84.43 & 92.73 & 93.65  & 48.70  & 50.97 \\
TCP  &  91.76 &  94.17  & 85.71  & 87.26 & 92.41 & 94.86  & 46.63  & 47.46 \\
\midrule
Ours (TRIP) & \textbf{95.84}* & \textbf{97.28}* & \textbf{90.23}*  & \textbf{91.39}* & \textbf{95.61}* & \textbf{96.87}* & \textbf{69.76}*  & \textbf{70.48}* \\
\bottomrule
\end{tabular}}
\label{tab:planning_zh}
\end{table}

\begin{table}[t!]
\caption{Experimental results of dialogue planning on the DuRecDial 2.0 dataset. Significant improvements over baseline models are marked with * (t-test, $p < 0.05$).}
\centering
\resizebox{0.7\linewidth}{!}{
\begin{tabular}{l|cc|cc|cc|cc}
\toprule
\multirow{3}{*}{\textbf{Model}} &  \multicolumn{4}{c|}{\textbf{Test-ID}} &  \multicolumn{4}{c}{\textbf{Test-OOD}} \\
\cmidrule{2-9}
 & \multicolumn{2}{c|}{\textbf{ Action}} &  \multicolumn{2}{c|}{\textbf{ Topic}} & \multicolumn{2}{c|}{\textbf{ Action}} &  \multicolumn{2}{c}{\textbf{ Topic}}\\
 & \textbf{F1} &  \textbf{Bi. F1}  & \textbf{F1} & \textbf{Bi. F1} & \textbf{F1} &  \textbf{Bi. F1}  & \textbf{F1} & \textbf{Bi. F1} \\
\midrule
MGCG & 90.26 &  92.47 &  74.93  & 79.24 & 82.30 & 87.25  & 36.03  & 42.00  \\
KERS & 90.33 &  91.54 &  77.85  & 80.35 & 84.21 & 86.39  & 34.20  & 37.85 \\
BERT & 91.68 &  92.37 &  79.21  & 81.22 & 92.23 & 94.19  & 46.55  & 52.12 \\
TCP  & 92.25 &  93.82 &  85.77  & 87.25 & 89.93 & 92.09  & 44.49  & 50.71 \\
\midrule
Ours (TRIP) & \textbf{94.49}* & \textbf{95.89}* & \textbf{91.83}*  & \textbf{93.51}* & \textbf{93.27}* & \textbf{95.18}* & \textbf{70.65}*  & \textbf{74.47}* \\
\bottomrule
\end{tabular}}
\label{tab:planning_en}
\end{table}

We observe similar trends in Table \ref{tab:dialogue_en} regarding automatic evaluation results on the DuRecDial 2.0 dataset. Both our prompt-based generation and plan-controlled generation methods outperform existing baseline models over most evaluation metrics. We note that all baseline models and our methods perform inferior to that on the DuRecDial dataset in terms of the goal success rate. It is because, in the DuRecDial 2.0 dataset, the domain knowledge triples grounding on each dialogue are noisier than that in the DuRecDial dataset, making it non-trivial for these models to distinguish the target topic and to generate the target topic in the utterance accordingly when necessary. Nonetheless, our methods still achieve better goal success rates, especially when evaluated on the out-of-domain (OOD) test set. Overall, experimental results reported in Table \ref{tab:dialogue_zh} and Table \ref{tab:dialogue_en} demonstrate that compared to existing methods, our proposed two variants are effective in generating more appropriate utterances on the end task of target-oriented dialogue generation.

\subsection{Evaluation Results of Dialogue Planning (RQ2)}

To validate the performance of dialogue action planning and topic planning, we compare our proposed TRIP model with existing dialogue planning models. The automatic evaluation results on the DuRecDial and DuRecDial 2.0 datasets are reported in Table \ref{tab:planning_zh} and \ref{tab:planning_en}, respectively. As shown in Table \ref{tab:planning_zh}, it is more difficult for all models to predict or generate dialogue topics correctly than dialogue actions because the total size of the topics is much larger than that of the actions in the dataset. For example, MGCG and KERS achieve comparable F1 and Bi. F1 scores on action planning while they perform much inferior on topic planning compared to other baseline models (i.e., BERT and TCP) that employ pre-trained language models. More obviously, we find that all models obtain much lower F1 and Bi. F1 scores in terms of topic planning when evaluated on the out-of-domain (OOD) test set. Since the target topics in the OOD test set are not allowed to appear in the train set, all models are challenging to capture the semantics of the target topics and predict or generate the target topics correctly. In contrast, our TRIP model achieves substantial improvements in both dialogue action planning and topic planning. Particularly, TRIP improves the topic F1 score from 70\%-80\% to over 90\% on the in-domain (ID) test set. It still maintains a much higher topic F1 score of 69.76\% on the challenging out-of-domain (OOD) test set. Similar trends are also observed in Table \ref{tab:planning_en} when all these methods are evaluated on the DuRecDial 2.0 dataset. We can conclude that our TRIP is able to plan a dialogue path consisting of more accurate dialogue actions and more reasonable topics. It is our effective dialogue planning that makes it possible to steer the system to lead the conversation toward the target proactively and smoothly.

\begin{table}[th!]
\caption{Ablation study results of our TRIP model on the DuRecDial dataset.}
\centering
\resizebox{0.68\linewidth}{!}{
\begin{tabular}{l|cc|cc|cc|cc}
\toprule
\multirow{3}{*}{\textbf{Model}} &  \multicolumn{4}{c|}{\textbf{Test-ID}} &  \multicolumn{4}{c}{\textbf{Test-OOD}} \\
\cmidrule{2-9}
 & \multicolumn{2}{c|}{\textbf{ Action}} &  \multicolumn{2}{c|}{\textbf{ Topic}} & \multicolumn{2}{c|}{\textbf{ Action}} &  \multicolumn{2}{c}{\textbf{ Topic}}\\
 & \textbf{F1} &  \textbf{Bi. F1}  & \textbf{F1} & \textbf{Bi. F1} & \textbf{F1} &  \textbf{Bi. F1}  & \textbf{F1} & \textbf{Bi. F1} \\
\midrule
TRIP (full) & 95.84 & 97.28 & 90.23 & 91.39 & 95.61 & 96.87 & 69.76  & 70.48  \\
\quad w/o $D^{F}$ & 93.71 & 95.55 &  86.26  & 87.40  & 92.09 & 93.17 & 46.52 & 50.22 \\
\quad w/o $D^{B}$ & 92.89 & 94.68 &  85.89  & 87.02  & 91.16 & 92.87 & 45.09 & 48.67 \\
\quad w/o $\mathcal{L}_{d}$ & 95.33 & 96.81 & 88.15  & 90.06 & 94.03 & 95.11 &  68.12 &  69.80 \\
\quad w/o $\mathcal{L}_{CL}$ & 95.45 & 96.90 & 88.76 &  90.13 & 93.15  & 94.09 & 67.30 & 68.22  \\
\quad w/o LC & 91.31 & 93.08 & 84.20 & 85.66 & 91.02 & 93.23 & 51.34 & 53.06 \\
\quad w/o BA & 92.06 & 94.35 & 85.46 & 86.89 & 91.15 & 93.20 & 52.49 & 54.13 \\
\bottomrule
\end{tabular}}
\label{tab:ablation}
\end{table}

\subsection{Ablation Study of TRIP (RQ3)}

To explore why our TRIP achieves superior performance in dialogue planning, we conducted an ablation study to verify the effectiveness of the modules and mechanisms proposed in TRIP. We focus on the following settings for ablation experiments: \textbf{(1)} without the forward decoder (w/o $D^{F}$), which denotes that only the backward decoder is employed to generate the dialogue path from the target turn to the present turn, followed by vanilla beam search decoding (the proposed target-constrained decoding algorithm is invalid in such a case); \textbf{(2)} without the backward decoder (w/o $D^{B}$), which denotes that only the forward decoder is employed to generate the dialogue path from the present turn to the target turn, followed by vanilla beam search decoding similarly; \textbf{(3)} without reducing the gap between backward-forward paths (w/o $\mathcal{L}_{d}$); \textbf{(4)} without the contrastive generation of targets (w/o $\mathcal{L}_{CL}$); \textbf{(5)} without the lexical constraints in the target-constrained decoding (w/o LC); \textbf{(6)} without the bidirectional agreement in the target-constrained decoding (w/o BA).

From the ablation study results shown in Table \ref{tab:ablation}, we observe that each module or mechanism contributes to dialogue planning. The performance of TRIP sharply dropped when removing either the backward decoder $D^{B}$ or the forward decoder $D^{F}$. In particular, the topic F1 score decreased from 69.76\% to 46.52\% (w/o $D^{F}$) and 45.09\% (w/o $D^{B}$) on the out-of-domain (OOD) test set. Such ablation results prove that our basic idea of employing two decoders for bidirectional planning is viable and effective. 
We also observe that the absence of $D^{B}$ performs worse than that of $D^{F}$. It is because $D^{B}$ directly takes the target as the beginning input of the decoder and generates the dialogue path in a target-to-present direction, which is of benefit to leverage the target-side information to guide planning more effectively. 
For the ablation results without $\mathcal{L}_{d}$ and $\mathcal{L}_{CL}$, both reducing the gap between backward-forward paths and contrastive generation of targets can benefit the model in planning as we expect. In terms of the target-constrained decoding, we find that the ultimate performance deteriorated rapidly when removing the lexical constraints (w/o LC) or bidirectional agreement (w/o BA), especially the topic F1 score decreased from 69.76\% to 51.34\% (w/o LC) and 52.49\% (w/o BA) on the OOD test set. It indicates that our target-constrained decoding performs a vital role in dialogue planning since it controls the model's attention to the target-side information during inference even when handling out-of-domain target topics.

\begin{figure*}[t!]
\centering
\subfigure[Impact of the hyperparameter $\lambda$]{
    \begin{minipage}[b]{0.4\textwidth}
	    \includegraphics[width=1\linewidth]{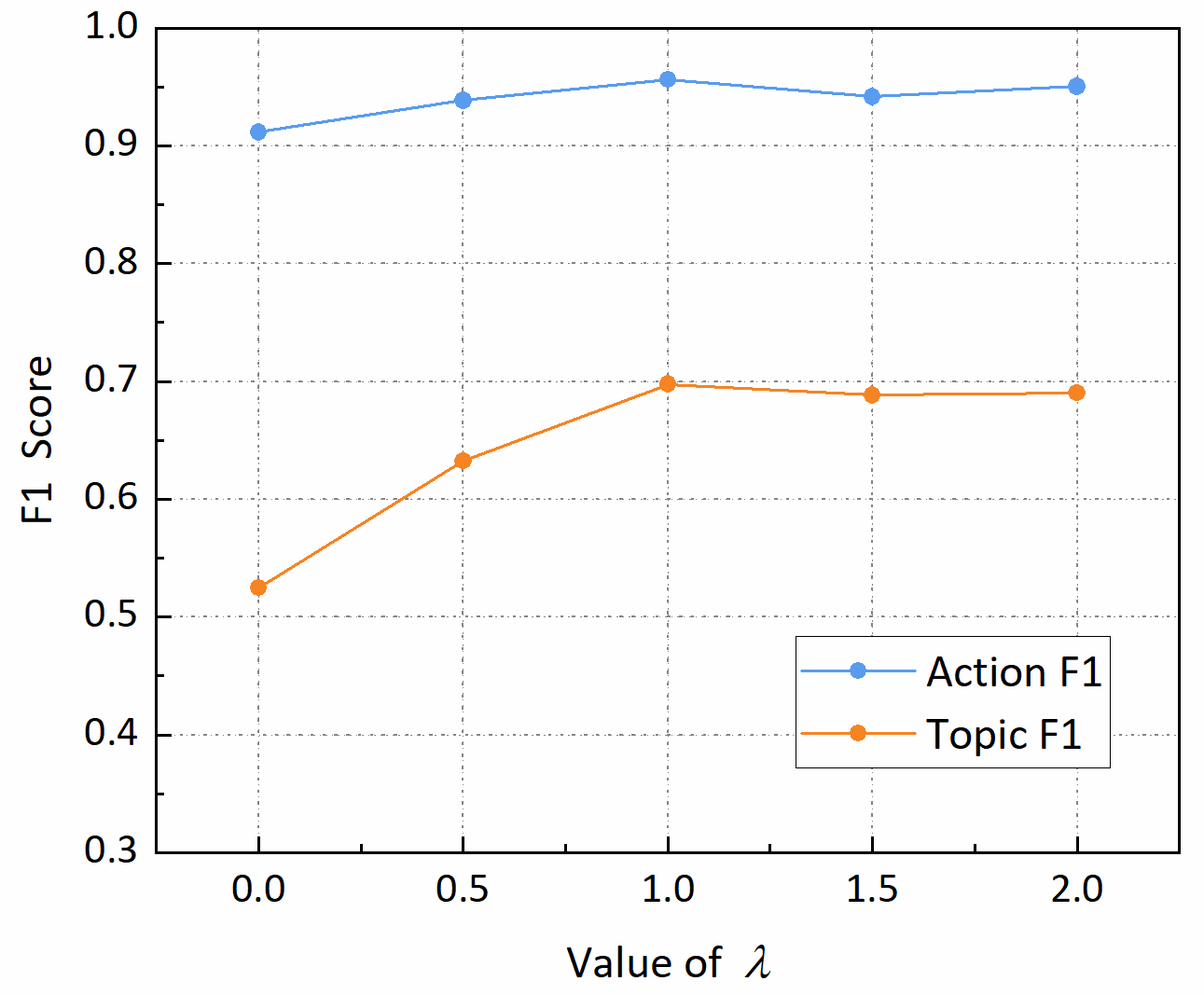}
    \end{minipage}
    \label{fig:parameter.a}
}
\subfigure[Impact of the step size $\alpha$]{
    \begin{minipage}[b]{0.44\textwidth}
	    \includegraphics[width=1\linewidth]{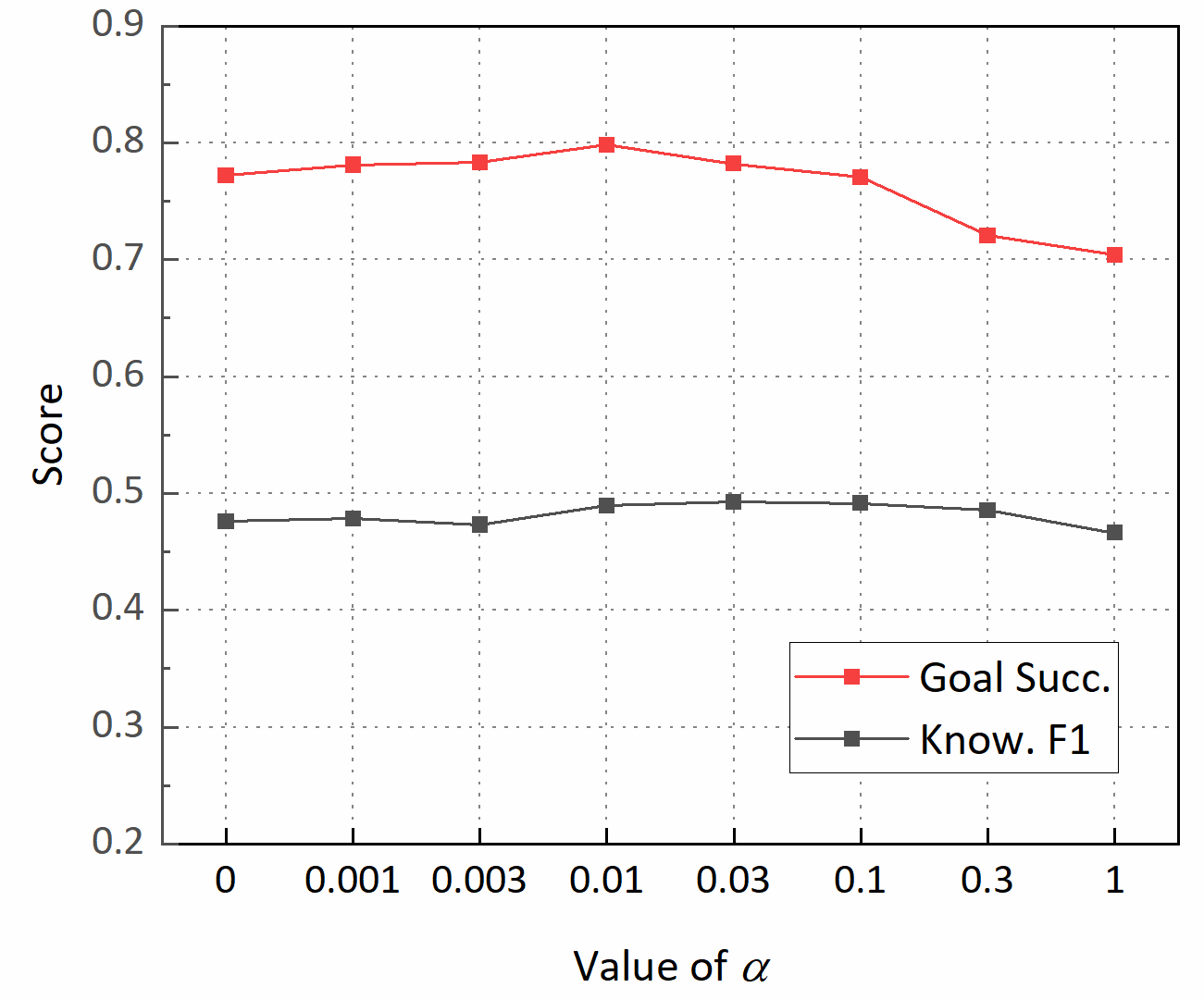}
    \end{minipage}
    \label{fig:parameter.b}
}
\caption{Quantitative results by varying the value of different parameters.}
\label{fig:parameter}
\end{figure*}

\subsection{Analysis of Parameters (RQ3)}
We quantitatively analyzed some critical parameters of our methods, including 1) the hyperparameter $\lambda$ that controls the weight of the bidirectional agreement reward in the planning stage and, 2) the step size $\alpha$ that controls the updating step in the plan-controlled dialogue generation.

\paragraph{\textbf{Impact of the hyperparameter $\lambda$}}
To investigate the impact of the hyperparameter $\lambda$ in the planning stage, we conducted target-constrained decoding by varying $\lambda$ in \{0, 0.5, 1.0, 1.5, 2.0\}. Experimental results are shown in Figure \ref{fig:parameter.a}. We observe that our model achieves the best action F1 and topic F1 scores when $\lambda=1.0$ and a smaller value of $\lambda$ results in lower action F1 and topic F1 scores. Particularly, the model performs much inferior without any reward of bidirectional agreement, i.e., $\lambda=0$, indicating that our target-constrained decoding with a bidirectional agreement is crucial in generating a more reasonable dialogue path.

\paragraph{\textbf{Impact of the step size $\alpha$}}
To investigate the impact of the step size $\alpha$ in the plan-controlled dialogue generation, we varied $\alpha$ by selecting its value in \{0, 0.001, 0.003, 0.01, 0.1, 0.3, 1.0\}. Experimental results are shown in Figure \ref{fig:parameter.b}. We observe that the step size $\alpha$ mainly affects the goal success rate while it has a slighter impact regarding the knowledge F1 score. If no updating step is performed during plan-controlled dialogue generation, i.e., $\alpha=0$, the dialogue generation model (i.e., LM $p(y)$) has no control of the output distribution, especially for those utterances that the target topics should explicitly appear. By default, we choose 0.01 as the most proper step size since neither a larger value nor a smaller one will bring any performance gain.

\begin{figure*}[t!]
\centering
\includegraphics[width=0.7\textwidth]{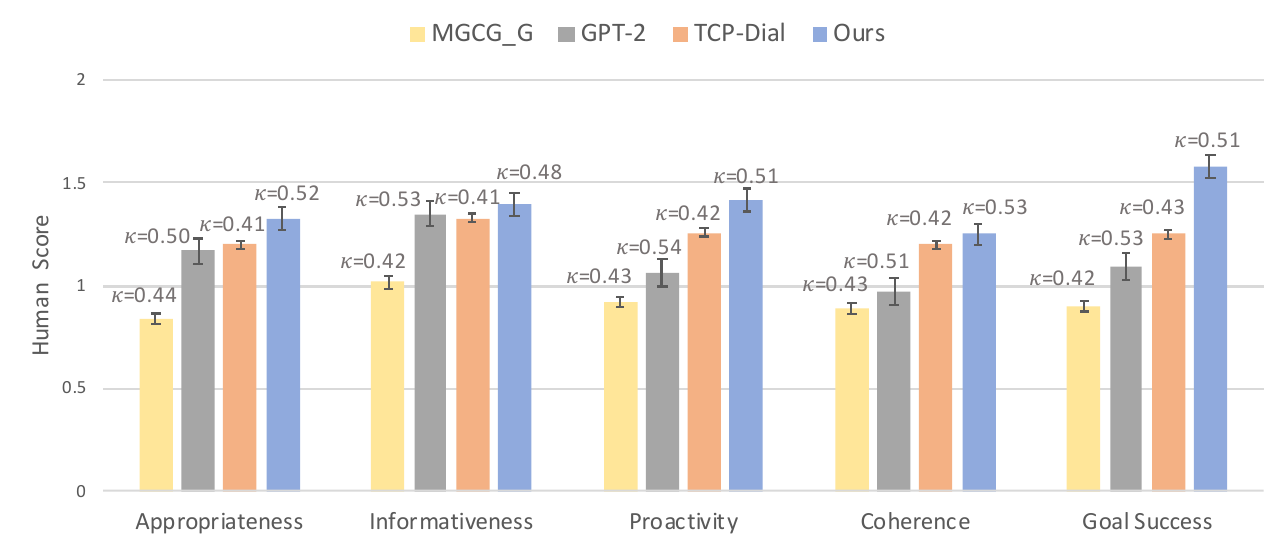}
\caption{Human evaluation results of different models. $\kappa$ denotes Fleiss's kappa.}
\label{fig:human_eval}
\end{figure*}

\subsection{Human Evaluation Results (RQ4)}

We selected several representative models for human evaluation, including MGCG\_G, GPT-2, TCP-Dial, and ours. The evaluation results are shown in Figure \ref{fig:human_eval}. The Fleiss's kappa scores are mainly distributed in [0.4, 0.6], denoting moderate inter-annotator agreement. For turn-level evaluation, we observe that GPT-2, TCP-Dial, and ours obtain comparable scores in informativeness since they utilize powerful pre-trained language models and thus can generate informative utterances. In terms of appropriateness, our method obtains the highest human score on average, demonstrating the ability to generate more appropriate system utterances in response to dialogue context. On the other hand, dialogue-level evaluation (i.e., proactivity, coherence, and goal success) is more challenging for all models because errors might be propagated as the dialogue goes on. We find that our method obtains better results on average compared to all baseline models. Notably, our method achieves the highest proactivity and goal success scores, indicating that our method is more likely to drive the dialogue to reach the target successfully.

\begin{figure*}[t!]
\centering
\subfigure[]{
    \begin{minipage}[b]{0.92\textwidth}
	    \includegraphics[width=1\linewidth]{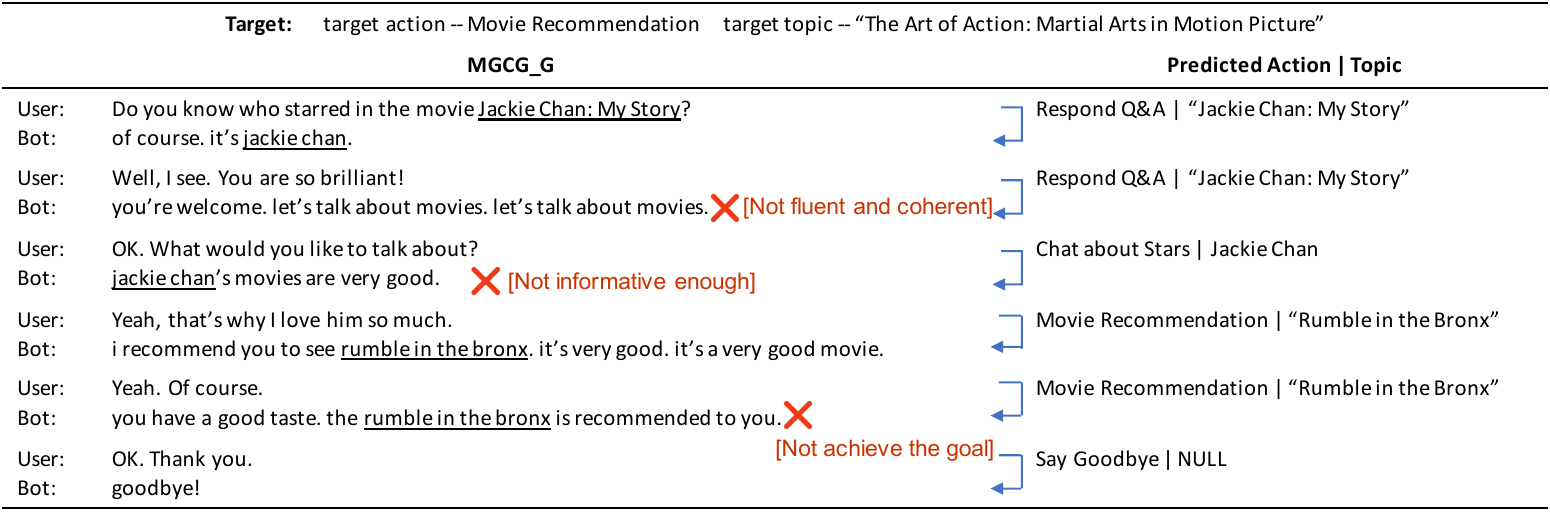}
    \end{minipage}
    \label{fig:case.a}
}
\subfigure[]{
    \begin{minipage}[b]{0.92\textwidth}
	    \includegraphics[width=1\linewidth]{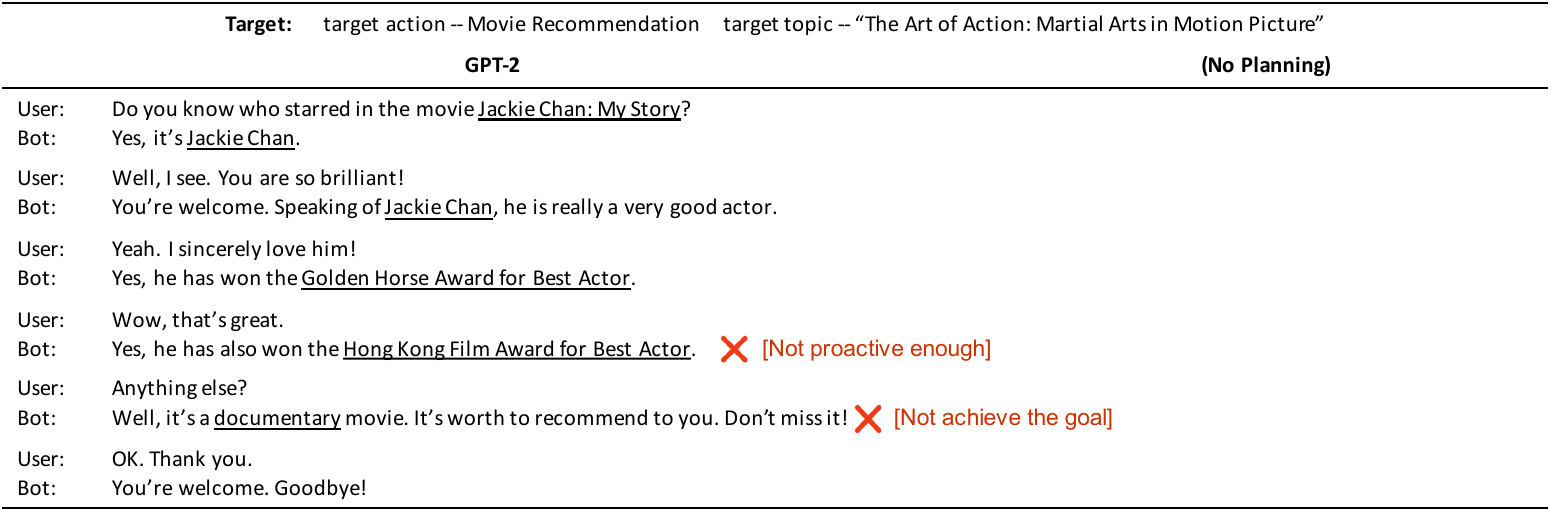}
    \end{minipage}
    \label{fig:case.b}
}
\subfigure[]{
    \begin{minipage}[b]{0.99\textwidth}
	    \includegraphics[width=1\linewidth]{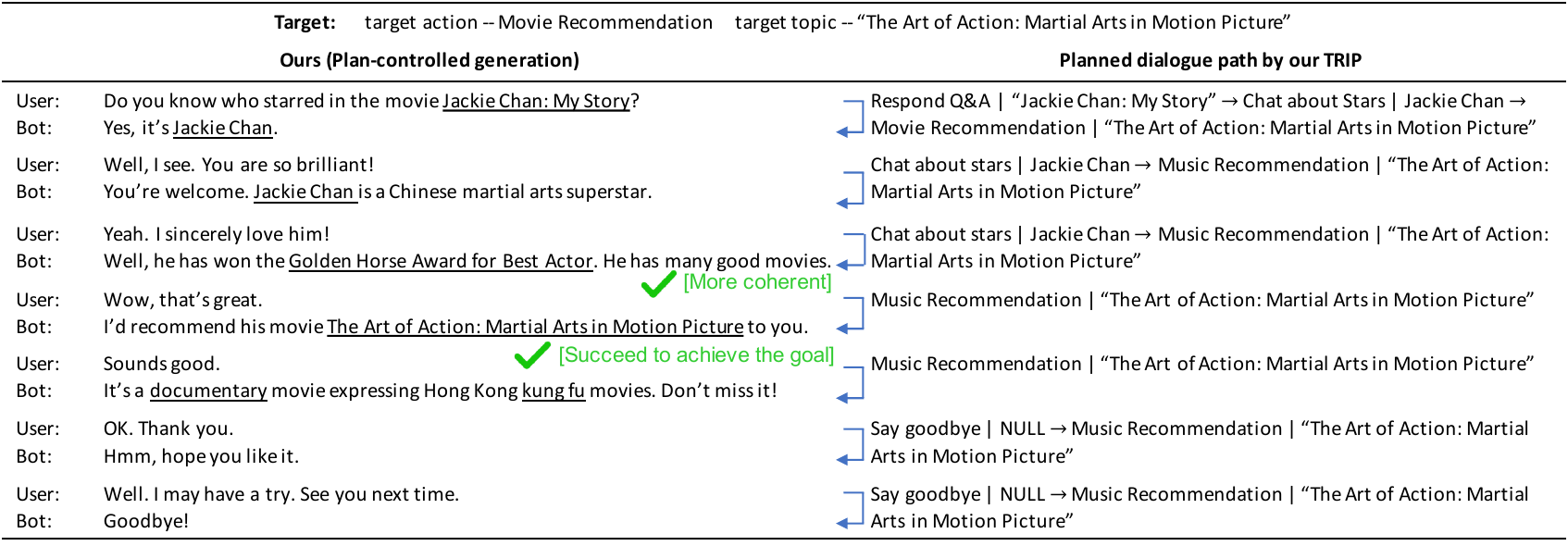}
    \end{minipage}
    \label{fig:case.c}
}
\caption{Illustrative cases from the dialogue-level human evaluation. The bot's utterances are generated by (a) MGCG\_G, (b) GPT-2, and (c) Ours (plan-controlled generation), respectively. The topics and topic-related attributes that also appear in the domain knowledge are marked with underlines.}
\label{fig:case}
\end{figure*}

\subsection{Case Study (RQ4)}

To illustrate the quality of different methods for target-oriented dialogue generation, we conducted some case studies. We selected the same target with the same initial dialogue context and investigated the generated utterances by three different models from dialogue-level human evaluation, including MGCG\_G, GPT-2, and ours (plan-controlled generation). Here, we show some generated cases in Figure \ref{fig:case}. As shown in Figure \ref{fig:case.a}, we observe that MGCG\_G is incapable of generating fluent and coherent utterances. Although MGCG\_G conducts planning first to predict the next dialogue action and topic, it fails to predict a correct topic when necessary, causing the model fails to achieve the target (i.e., recommend the movie ``\textit{The Art of Action: Martial Arts in Motion Picture}'') at the end of the dialogue. For the case of GPT-2 shown in Figure \ref{fig:case.b}, we find that GPT-2 is able to generate more fluent and informative utterances in general. However, it fails to achieve the target since it has no dialogue planning, making it not proactive enough to lead the dialogue towards the pre-determined target. In such cases, GPT-2 is not effective to generate the target topic as the dialogue goes on. In contrast, the case shown in Figure \ref{fig:case.c} demonstrates that our TRIP model can plan a dialogue path with reasonable actions and appropriate topics that outlines how to achieve the target step by step. With the guidance of the planned path, our plan-controlled generation method can know when and what to talk about to move the dialogue forward proactively. More importantly, our method succeeds in achieving the target since our TRIP plans a correct topic (i.e., the target topic ``\textit{The Art of Action: Martial Arts in Motion Picture}'') when appropriate.

\subsection{Additional Discussions (RQ4)}
According to the human evaluation results and case study, our proposed methods effectively plan reasonable dialogue paths to guide dialogue generation. The advantages of such a pipelined framework are: (1) It provides our model with better explanations because each planned dialogue path tells the dialogue generation model how to achieve the target step by step with specific actions and essential topics. (2) It is controllable for the end task of target-oriented dialogue generation. Our methods divide the complicated end task into two stages, making it more flexible to improve the overall performance stage by stage. Therefore, our methods are more practical and can be extended to real-world applications.
After analyzing those cases with low human evaluation scores, we also identify some limitations and discuss the potential solutions: (1) Our pipelined framework has error propagation, which might be a typical issue of most existing pipelined methods. We find that the performance of dialogue generation is prone to drop once our TRIP model fails to plan a dialogue path appropriately. We intend to alleviate this issue by introducing some techniques in the cascaded generation, such as noisy channel models \cite{shannon1948mathematical,liu2021pretraining}. (2) Our plan-guided dialogue generation method is still not robust enough. Although we have achieved significant good planning results with a large margin compared to baseline models on both datasets, we observe that the performance gain in terms of the goal success rate is much less prominent on one dataset than on another. One possible direction is to study how to improve dialogue generation with adaptive control when it is the turn with the target action and the target topic.

\section{Conclusion and Future Work}
In this work, we explore the task of target-oriented proactive dialogue and focus on effective dialogue planning for dialogue generation. We propose a novel target-constrained bidirectional planning (TRIP) approach to plan dialogue paths from both backward and forward directions. Our TRIP formulates planning as a generation task and bidirectionally generates dialogue paths consisting of reasonable actions and appropriate topics. To better control path generation, we devise a novel target-constrained decoding algorithm to achieve bidirectional agreement. We adopt the planned dialogue paths to guide dialogue generation in a pipeline manner, with two explored variants: prompt-based generation and plan-controlled generation. Experimental results on two re-purposed datasets show that the proposed methods achieve state-of-the-art performance on all sub-tasks. Extensive analysis and discussions demonstrate the advantages of our methods. 

We observe that the emergence of large language models (LLMs) \cite{openai2022chatgpt,touvron2023llama,touvron2023llama2} has unprecedentedly boosted the research field of dialogue systems. LLMs will generally perform better for dialogue generation in terms of some aspects, such as fluency, informativeness, and human likeness. However, for the target-oriented proactive dialogue generation task, more critical dimensions should be considered, including proactivity, coherence, and target achievement success rate. Our work shows that dialogue planning plays a vital role in improving dialogue generation performance in these dimensions. 
Recent studies \cite{valmeekam2022large,valmeekam2023planning} indicate that the planning capabilities of LLMs are still far from that of humans. In the future, we intend to incorporate our proposed bidirectional approach based on LLMs for dialogue planning and generation since our methods are model-agnostic to backbone models. We are also interested in empowering the planning capabilities of LLMs to solve other complex tasks.

\begin{acks}
This work was supported by the Research Grants Council of Hong Kong (15207122, 15207920, 15207821, 15204018, 15213323) and National Natural Science Foundation of China (62076212). It was also supported in part by PolyU internal grants (ZVQ0, ZVVX).

\end{acks}

\bibliographystyle{ACM-Reference-Format}
\bibliography{sample-base}


\end{document}